\documentclass[10pt,conference,letterpaper,final]{IEEEtran}

\usepackage[T1]{fontenc}
\usepackage[utf8]{inputenc}

\usepackage[pdftex]{graphicx}
\usepackage{amsmath}
\usepackage{amssymb}
\usepackage{array}

\usepackage[hidelinks]{hyperref}
\usepackage{url}

\ifCLASSOPTIONcompsoc
  \usepackage[caption=false,font=normalsize,labelfont=sf,textfont=sf]{subfig}
\else
  \usepackage[caption=false,font=footnotesize]{subfig}
\fi

\usepackage{cleveref}
\crefname{figure}{fig.}{figs.}
\Crefname{figure}{Fig.}{Figs.}
\crefname{equation}{}{}
\Crefname{equation}{Equation}{Equations} 
\crefname{section}{Section}{Sections} 
\crefname{subsection}{Section}{Sections} 
\Crefname{subsection}{Section}{Sections} 

\usepackage{blindtext}
\usepackage[detect-all]{siunitx} 
\usepackage{mathtools}
\usepackage[olditem,oldenum]{paralist}
\DeclareSIUnit\event{Event}

\usepackage[permil]{overpic}
\usepackage{tikz}
\usetikzlibrary{patterns}

\usepackage{algorithmic}

\usepackage{textcomp}

\usepackage[textsize=tiny,obeyFinal]{todonotes}

\usepackage{microtype}

\usepackage{booktabs}

\usepackage{fixltx2e}
\usepackage{dblfloatfix}
\usepackage[noadjust]{cite}

\usepackage{ifdraft}
\ifoptionfinal{}{%
  \usepackage[switch]{lineno}
  \linenumbers
}

\usetikzlibrary{backgrounds}

\renewcommand{\arraystretch}{1.3}

\newcommand\newreplacement[2]{\def#1/{#2}}

\newreplacement{\itl}{in-the-loop}
\newreplacement{\Itl}{In-the-loop}
\newreplacement{\ITL}{In-The-Loop}
\newreplacement{\itlnodashes}{in the loop}
\newreplacement{\Itlnodashes}{In the loop}
\newreplacement{\ITLnodashes}{In The Loop}

\newreplacement{\TFnet}{artificial network}

\newreplacement{\NMnet}{neuromorphic network}

\newreplacement{\BSS}{BrainScaleS}
\newreplacement{\BSSS}{BrainScaleS system}
\newreplacement{\BSSWSS}{BrainScaleS wafer-scale system}
\newreplacement{\BSSWM}{BrainScaleS wafer module}

\newcommand\asymunc[4]{\ensuremath{#1\,\substack{+#2\\-#3}}\,#4}

\newcommand\class[1]{``#1''}

\makeatletter
\g@addto@macro\bfseries{\boldmath}
\makeatother

\begin{document}

\bstctlcite{IEEEexample:BSTcontrol}




\newcommand{\bs}[1]{\boldsymbol{#1}}
\newcommand{\minisection}[1]{\paragraph{\emph{#1}}}
\newcommand{\paramtype}[2]{\mathsf{#1_{#2}}}
\newcommand{\sidenote}[1]{\marginnote{\emph{#1}}}
\newcommand{\tb}[1]{\textbf{#1}}
\newcommand{\ti}[1]{\textit{#1}}
\newcommand{\todoin}[1]{\todo[inline]{#1}}


\newcommand{\Bigcap}{\bigcap\limits}
\newcommand{\DKL}[2]{D_\mathrm{KL}\left(#1\parallel#2\right)}
\newcommand{\DKLnorm}[2]{D_\mathrm{KL}^\mathrm{norm}\left(#1\parallel#2\right)}
\newcommand{\Expect}[1]{E\left[#1\right]}
\newcommand{\expect}[1]{\left\langle#1\right\rangle}
\newcommand{\Int}{\int\limits}
\newcommand{\Lim}{\lim\limits}
\newcommand{\Prod}{\prod\limits}
\newcommand{\Sum}{\sum\limits}
\newcommand{\var}[1]{\Var\left[#1\right]}


\newcommand{\ci}[3]{#1 {\perp\!\!\!\perp} #2 \; | \; #3}
\newcommand{\nci}[3]{#1 \centernot{\perp\!\!\!\perp} #2 \; | \; #3}


\newcommand{\abstr}{\mathrm{abstr}\xspace}
\newcommand{\bmErev}{\bm{E}^\mathrm{rev}\xspace}
\newcommand{\bmtausyn}{\bm\tau^\mathrm{syn}\xspace}
\newcommand{\bmueff}{\bm{u}_\mathrm{eff}\xspace}
\newcommand{\boxfct}{\mathrm{box}\xspace}
\newcommand{\Ca}{{\mathrm{Ca}^{++}}\xspace}
\newcommand{\CC}{\rho\xspace}
\newcommand{\Cl}{{\mathrm{Cl}^-}\xspace}
\newcommand{\Cm}{{C_\mathrm{m}}\xspace}
\newcommand{\const}{{\mathrm{const}}\xspace}
\newcommand{\CVISI}{{\mathrm{CV}_\mathrm{ISI}}\xspace}
\newcommand{\CVrate}{{\mathrm{CV}_\mathrm{rate}}\xspace}
\newcommand{\ddt}{{\frac{d}{dt}}\xspace}
\newcommand{\DeltaT}{{\Delta_\mathrm{T}}\xspace}
\newcommand{\DKLsolo}{{D_\mathrm{KL}}\xspace}
\newcommand{\El}{{E_\mathrm{l}}\xspace}
\newcommand{\Eqn}{\mathrm{Eqn.}\xspace}
\newcommand{\Eqns}{\mathrm{Eqns.}\xspace}
\newcommand{\Er}{{E_\mathrm{r}}\xspace}
\newcommand{\Erev}{E^\mathrm{rev}\xspace}
\newcommand{\Ereve}{{E^\mathrm{rev}_\mathrm{e}}\xspace}
\newcommand{\Erevi}{{E^\mathrm{rev}_\mathrm{i}}\xspace}
\newcommand{\ET}{{E_\mathrm{T}}\xspace}
\newcommand{\gext}{{g_\mathrm{ext}}\xspace}
\newcommand{\gimax}{{g_i^\mathrm{max}}\xspace}
\newcommand{\gl}{{g_\mathrm{l}}\xspace}
\newcommand{\fsyn}{{f^\mathrm{syn}}\xspace}
\newcommand{\gsyn}{g^\mathrm{syn}\xspace}
\newcommand{\gsyne}{{g^\mathrm{syn}_\mathrm{e}}\xspace}
\newcommand{\gsyni}{{g^\mathrm{syn}_\mathrm{i}}\xspace}
\newcommand{\gtot}{{g^\mathrm{tot}}\xspace}
\newcommand{\icb}{\texttt{icb}}
\newcommand{\iext}{{i^\mathrm{ext}}\xspace}
\newcommand{\Iext}{I^\mathrm{ext}\xspace}
\newcommand{\ifmath}{{\mathrm{if}}\xspace}
\newcommand{\Inoise}{I^\mathrm{noise}}
\newcommand{\ipi}{{}^1p_1\xspace}
\newcommand{\irc}{{i^\mathrm{RC}}\xspace}
\newcommand{\Irec}{I^\mathrm{rec}}
\newcommand{\Iref}{{I_\mathrm{ref}}\xspace}
\newcommand{\isyn}{i^\mathrm{syn}\xspace}
\newcommand{\Isyn}{I^\mathrm{syn}\xspace}
\newcommand{\Jsyn}{J^\mathrm{syn}\xspace}
\newcommand{\lambdam}{{\lambda_\mathrm{m}}\xspace}
\newcommand{\Mexc}{M_\mathrm{exc}\xspace}
\newcommand{\Minh}{M_\mathrm{inh}\xspace}
\newcommand{\Na}{{\mathrm{Na}^+}\xspace}
\newcommand{\NBAS}{{N_\mathrm{BAS}}\xspace}
\newcommand{\nne}{{\mathrm{ne}}\xspace}
\newcommand{\NHC}{{N_\mathrm{HC}}\xspace}
\newcommand{\NMC}{{N_\mathrm{MC}}\xspace}
\newcommand{\non}{{\mathrm{\setminus}}\xspace}
\newcommand{\nonk}{{\non k}\xspace}
\newcommand{\NPYR}{{N_\mathrm{PYR}}\xspace}
\newcommand{\nRS}{{n_\mathrm{RS}}\xspace}
\newcommand{\nFS}{{n_\mathrm{FS}}\xspace}
\newcommand{\nusyn}{\nu^\mathrm{syn}\xspace}
\newcommand{\NRSNP}{{N_\mathrm{RSNP}}\xspace}
\newcommand{\otherwise}{\mathrm{otherwise}\xspace}
\newcommand{\pa}{\mathrm{\textbf{pa}}\xspace}
\newcommand{\pflip}{p_\mathrm{flip}\xspace}
\newcommand{\poo}{p_{00}\xspace}
\newcommand{\poi}{p_{01}\xspace}
\newcommand{\pio}{p_{10}\xspace}
\newcommand{\pii}{p_{11}\xspace}
\newcommand{\PSP}{\mathrm{PSP}\xspace}
\newcommand{\pspike}{p_\mathrm{spike}\xspace}
\newcommand{\rl}{{r_\mathrm{l}}\xspace}
\newcommand{\SU}{\tilde I\xspace}
\newcommand{\taubk}{\overline{\tau^\mathrm{b}_k}}
\newcommand{\taudecay}{{\tau_\mathrm{decay}}\xspace}
\newcommand{\taueff}{{\tau_\mathrm{eff}}\xspace}
\newcommand{\taufacil}{{\tau_\mathrm{facil}}\xspace}
\newcommand{\taufall}{{\tau_\mathrm{fall}}\xspace}
\newcommand{\tauinact}{{\tau_\mathrm{inact}}\xspace}
\newcommand{\taum}{{\tau_\mathrm{m}}\xspace}
\newcommand{\tauon}{{\tau_\mathrm{on}}\xspace}
\newcommand{\tauON}{{\tau_\mathrm{ON}}\xspace}
\newcommand{\taurise}{{\tau_\mathrm{rise}}\xspace}
\newcommand{\taurec}{{\tau_\mathrm{rec}}\xspace}
\newcommand{\tauref}{{\tau_\mathrm{ref}}\xspace}
\newcommand{\taustdp}{\tau^\mathrm{STDP}\xspace}
\newcommand{\tausyn}{\tau^\mathrm{syn}\xspace}
\newcommand{\tausyne}{{\tau^\mathrm{syn}_\mathrm{e}}\xspace}
\newcommand{\tausyni}{{\tau^\mathrm{syn}_\mathrm{i}}\xspace}
\newcommand{\tauw}{{\tau_\mathrm{w}}\xspace}
\newcommand{\textmax}{{\mathrm{max}}\xspace}
\newcommand{\textmin}{{\mathrm{min}}\xspace}
\newcommand{\thetaeff}{{\vartheta_\mathrm{eff}}\xspace}
\newcommand{\trise}{{t_\mathrm{rise}}\xspace}
\newcommand{\tfall}{{t_\mathrm{fall}}\xspace}
\newcommand{\tot}{\mathrm{tot}\xspace}
\newcommand{\tspike}{{t_\mathrm{spike}}\xspace}
\newcommand{\ueff}{{u_\mathrm{eff}}\xspace}
\newcommand{\unif}{{\mathrm{unif}}\xspace}
\newcommand{\ureset}{{u_\mathrm{reset}}\xspace}
\newcommand{\USE}{{U_\mathrm{SE}}\xspace}
\newcommand{\uthr}{{u_\mathrm{thr}}\xspace}
\newcommand{\Vm}{{V_\mathrm{m}}\xspace}
\newcommand{\Vrest}{{V_\mathrm{rest}}\xspace}
\newcommand{\Vspike}{{V_\mathrm{spike}}\xspace}
\newcommand{\Vth}{{V_\mathrm{th}}\xspace}
\newcommand{\Vthresh}{\ET}
\newcommand{\wsyn}{w^\mathrm{syn}\xspace}
\newcommand{\zpi}{{}^2p_1\xspace}


\newcommand{\COBA}{\mathrm{COBA}\xspace}
\newcommand{\Cov}{\mathrm{Cov}\xspace}
\newcommand{\CUBA}{\mathrm{CUBA}\xspace}
\newcommand{\erf}{\mathrm{erf}\xspace}
\newcommand{\for}{\mathrm{for}\xspace}
\newcommand{\sgn}{\mathrm{sgn}\xspace}
\newcommand{\Var}{\mathrm{Var}\xspace}


\newcommand{\aEIFcurrexp}{\mbox{\texttt{aEIF\_curr\_exp}}\xspace}
\newcommand{\IFcondalpha}{\mbox{\texttt{IF\_cond\_alpha}}\xspace}
\newcommand{\NEST}{\mbox{\texttt{NEST}}\xspace}
\newcommand{\Neuron}{\mbox{\texttt{Neuron}}\xspace}
\newcommand{\PyNN}{\mbox{\texttt{PyNN}}\xspace}


\newcommand{\tagarray}{\mbox{}\refstepcounter{equation}$(\theequation)$}
\newenvironment{texttab}[1]
    {\vspace{-10pt}
     \tabulinesep=7pt
     \begin{center}
         \begin{tabu} to 1.013\textwidth {#1}}
        {\end{tabu}
    \end{center}}

\def\layersep{2.5cm} 
\def\neuronsep{1.2} 
\tikzstyle{neuron}=[circle,fill=black!25,minimum size=21pt,inner sep=0pt]
\tikzstyle{visible neuron}=[neuron, fill=green!50]
\tikzstyle{hidden neuron}=[neuron, fill=orange!75]

\title{Neuromorphic Hardware In The Loop:\\Training a Deep Spiking Network on the BrainScaleS Wafer-Scale System}

\author{%
  \IEEEauthorblockN{%
    Sebastian~Schmitt\IEEEauthorrefmark{2}\enspace
    Johann~Kl\"ahn\IEEEauthorrefmark{2}\enspace
    Guillaume~Bellec\IEEEauthorrefmark{4}\enspace
    Andreas~Gr\"{u}bl\IEEEauthorrefmark{2}\enspace
    Maurice~G\"{u}ttler\IEEEauthorrefmark{2}\enspace
    \\
    Andreas~Hartel\IEEEauthorrefmark{2}\enspace
    Stephan~Hartmann\IEEEauthorrefmark{3}\enspace
    Dan~Husmann\IEEEauthorrefmark{2}\enspace
    Kai~Husmann\IEEEauthorrefmark{2}\enspace
    Sebastian~Jeltsch\IEEEauthorrefmark{2}\enspace
    \\
    Vitali~Karasenko\IEEEauthorrefmark{2}\enspace
    Mitja~Kleider\IEEEauthorrefmark{2}\enspace
    Christoph~Koke\IEEEauthorrefmark{2}\enspace
    Alexander~Kononov\IEEEauthorrefmark{2}\enspace
    Christian~Mauch\IEEEauthorrefmark{2}\enspace
    \\
    Eric~M\"{u}ller\IEEEauthorrefmark{2}\enspace
    Paul~M\"{u}ller\IEEEauthorrefmark{2}\enspace
    Johannes~Partzsch\IEEEauthorrefmark{3}\enspace
    Mihai~A.~Petrovici\IEEEauthorrefmark{2}\IEEEauthorrefmark{6}\enspace
    Stefan~Schiefer\IEEEauthorrefmark{3}\enspace
    \\
    Stefan~Scholze\IEEEauthorrefmark{3}\enspace
    Vasilis~Thanasoulis\IEEEauthorrefmark{3}\enspace
    Bernhard~Vogginger\IEEEauthorrefmark{3}\enspace
    Robert~Legenstein\IEEEauthorrefmark{4}\enspace
    \\
    Wolfgang~Maass\IEEEauthorrefmark{4}\enspace
    Christian~Mayr\IEEEauthorrefmark{3}\enspace
    René~Schüffny\IEEEauthorrefmark{3}\enspace
    Johannes~Schemmel\IEEEauthorrefmark{2}\enspace
    Karlheinz~Meier\IEEEauthorrefmark{2}
  }\\
  {\footnotesize
  \bgroup
  \def\arraystretch{1.0}
  \tt
  \begin{tabular}{c}
  \{sschmitt,\,kljohann,\,agruebl,\,gguettle,\,ahartel,\,husmann,\,khusmann,\,sjeltsch,\,vkarasen,\,mkleider,\,\\
  koke,\,akononov,\,cmauch,\,mueller,\,pmueller,\,mpedro,\,schemmel,\,meierk\}@kip.uni-heidelberg.de
  \vspace{0.15cm}
  \\
  \{stephan.hartmann,\,johannes.partzsch,\,stefan.schiefer,\,stefan.scholze,\,\\
  vasileios.thanasoulis,\,bernhard.vogginger,\,christian.mayr,\,rene.schueffny\}@tu-dresden.de
  \vspace{0.15cm}
  \\
  \{guillaume,\,robert.legenstein,\,maass\}@igi.tugraz.at
  \end{tabular}
  \egroup
  }
  \\
  \\
  \IEEEauthorblockA{
    \IEEEauthorrefmark{2}Heidelberg University, Kirchhoff-Institute for Physics, Im Neuenheimer Feld 227, D-69120 Heidelberg\\
    \IEEEauthorrefmark{3}Technische Universität Dresden, Chair for Highly-Parallel VLSI-Systems and Neuromorphic Circuits, D-01062 Dresden\\
    \IEEEauthorrefmark{4}Graz University of Technology, Institute for Theoretical Computer Science, %
    A-8010 Graz\\
    \IEEEauthorrefmark{6}University of Bern, Department of Physiology, Bühlplatz 5, CH-3012 Bern
  }
}

\hypersetup{pdfinfo={
  Author={Sebastian Schmitt, Johann Kl\"ahn, Guillaume Bellec, Andreas Gr\"{u}bl, Maurice G\"{u}ttler, Andreas Hartel, Stephan Hartmann, Dan Husmann, Kai Husmann, Sebastian Jeltsch, Vitali Karasenko, Mitja Kleider, Christoph Koke, Alexander Kononov, Christian Mauch, Eric M\"{u}ller, Paul M\"{u}ller, Johannes Partzsch, Mihai A. Petrovici, Bernhard Vogginger, Stefan Schiefer, Stefan Scholze, Vasilis Thanasoulis, Robert Legenstein, Wolfgang Maass, Christian Mayr, Johannes Schemmel, Karlheinz Meier},
  Title={Neuromorphic Hardware In The Loop: Training a Deep Spiking Network on the BrainScaleS Wafer-Scale System},
  Keywords={Deep Neural Networks, DNN, neuromorphic, accelerated, ReLU, rectified linear unit, PyNN, tensorflow, spiking, wafer-scale}
}
}

\maketitle

\begin{abstract}
Emulating spiking neural networks on analog neuromorphic hardware offers several advantages over simulating them on conventional computers, particularly in terms of speed and energy consumption.
However, this usually comes at the cost of reduced control over the dynamics of the emulated networks.
In this paper, we demonstrate how iterative training of a hardware-emulated network can compensate for anomalies induced by the analog substrate.
We first convert a deep neural network trained in software to a spiking network on the \BSS/ wafer-scale neuromorphic system, thereby enabling an acceleration factor of \num{10000} compared to the biological time domain.
This mapping is followed by the \itl/ training, where in each training step, the network activity is first recorded in hardware and then used to compute the parameter updates in software via backpropagation.
An essential finding is that the parameter updates do not have to be precise, but only need to approximately follow the correct gradient, which simplifies the computation of updates.
Using this approach, after only several tens of iterations, the spiking network shows an accuracy close to the ideal software-emulated prototype.
The presented techniques show that deep spiking networks emulated on analog neuromorphic devices can attain good computational performance despite the inherent variations of the analog substrate.

\end{abstract}

\IEEEpeerreviewmaketitle

\section{Introduction}

\begin{figure}[t]
  \centering
  \includegraphics[width=\linewidth]{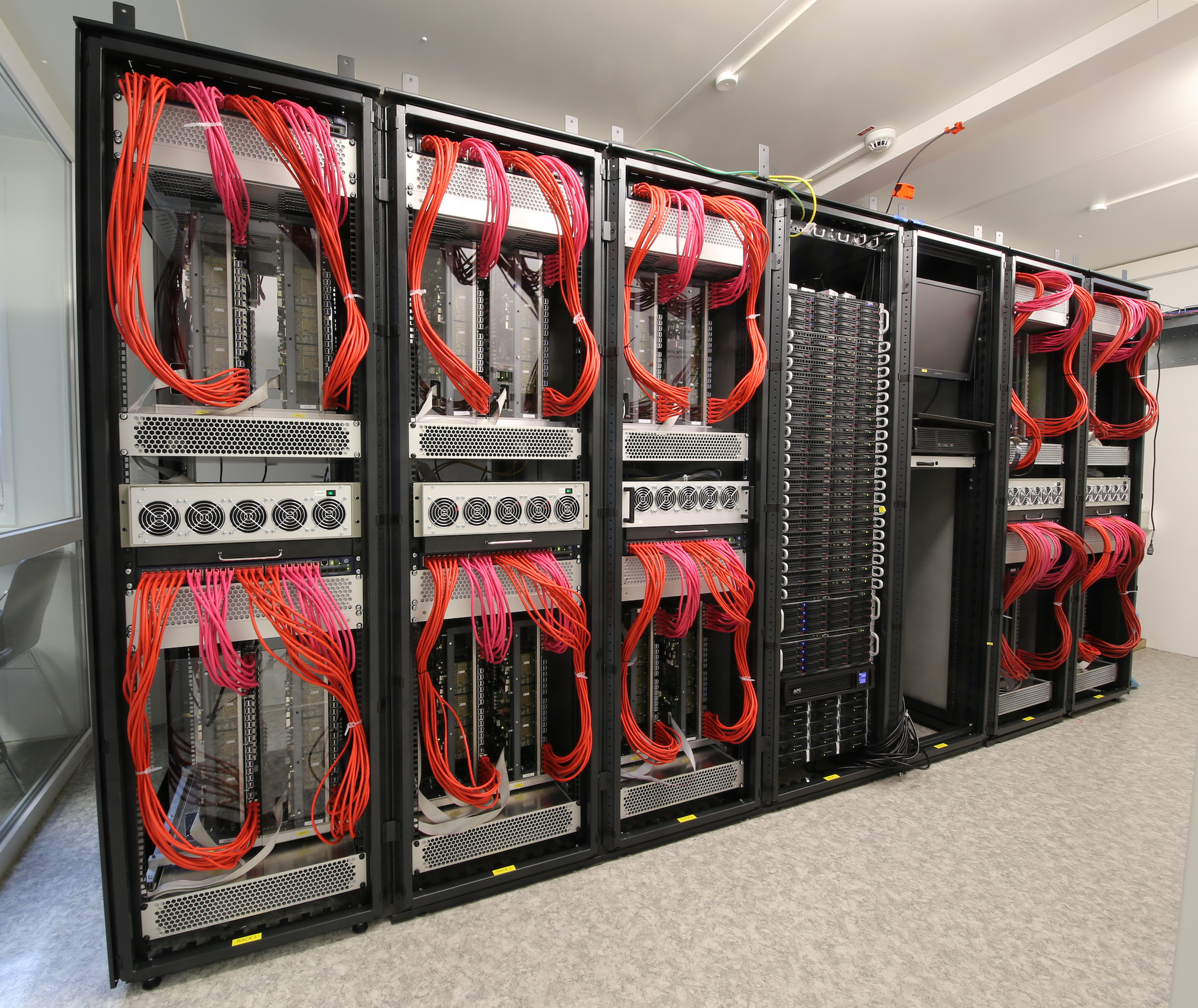}
  \caption{%
    The \BSSS/ as it is currently installed consisting of five cabinets, each containing four neuromorphic wafer-scale systems.
    Upstream connectivity to the control cluster is provided by the prominent red cables, each communicating at Gigabit speed.
    This enables fast system configuration and high-throughput spike in- and output.
    An additional rack hosts the support infrastructure comprising power supplies, servers, the control cluster,
    and network equipment.%
  }
  \label{fig:machineroom}
\end{figure}

Recently, artificial neural networks (ANNs) have emerged as the dominant machine learning paradigm for many pattern recognition problems~\cite{LeCun2015}.
Although ANNs are to some extent inspired by the architecture of biological neuronal networks, they differ significantly from their biological counterpart in many respects.
First, while the computation in biological neurons is performed through analog voltages in continuous time, ANNs are typically implemented on digital hardware and thus operate in discretized time.
Second, while the communication between neurons in an ANN is based on high-precision
arithmetic and computed in discrete time steps, communication in biological neuronal networks is largely based on stereotypically shaped all-or-none voltage events in continuous time.
These events are called action potentials or spikes.
In recent years, several large-scale
analog neuromorphic computing platforms have been developed~\cite{furber2016-large-scale-nm-system} that better match these features of biological neural networks.
Due to their low power consumption and speedup compared to simulations run on conventional architectures, these systems are promising precursors for computing devices that can rival the computational capabilities and energy efficiency of the human brain.

While spiking neural networks are in principle able to emulate any ANN~\cite{maass1997fast}, it has been unclear whether neuromorphic hardware can be efficiently used to implement contemporary deep ANNs.
One obstacle has been the lack of adequate training procedures.
ANNs are typically trained by backpropagation, a learning algorithm that propagates high-precision errors through the layers of the network.
Recently the successful training of neural networks was demonstrated on the TrueNorth chip, a fully digital spike-based neuromorphic design~\cite{Esser20092016}.
More specifically, performance on machine-learning benchmarks is not impaired by their hardware quantization constraints if, at each training step, the errors are computed with quantized parameters and binarized activations, before backpropagating with full precision.
This advance however left the question open whether a similar strategy could be used for analog neuromorphic systems.
Since TrueNorth is fully digital, an exact software model is available.
Therefore, each parameter, neuron activations, and the corresponding gradients are available or can be appropriately approximated at any point in time during training.
In contrast, the neural circuits on analog hardware are not as precisely controllable, making an exact mapping between the hardware and software domains challenging.

In this work, we demonstrate the successful training of an analog neuromorphic system configured to implement a deep neural architecture.
The system we used is the \BSSWSS/, a mixed-signal neuromorphic architecture that features analog neuromorphic circuits with digital, event-based communication.
We implemented a training procedure similar to~\cite{Esser20092016}, but used only a coarse software model to approximate its behavior.
We show that, nevertheless, the backpropagation algorithm is capable to adapt the synaptic parameters of the neuromorphic network quite effectively when running the training with the hardware \itlnodashes/.
Similar approaches have already been used, in the context of various network architectures, for smaller analog neuromorphic platforms, such as the HAGEN~\cite{hohmann_ijcnn04,fieres06convolutional} and Spikey~\cite{pfeil2013six} chips.

For the parameter updates, we used the recorded activity of the neuromorphic system, but computed the corresponding gradients using the parameters of the ANN.
This adaptation was possible in spite of the fact that the algorithm had no explicit knowledge about exact parameter values of the neurons and synapses in the \BSSS/.

The remainder of the article is structured as follows.
In \cref{sec:bss}, we describe the \BSS/ neuromorphic platform and discuss the extent of parameter variability in this system.
Starting from a simple approximate software model, \cref{subsec:initialtraining}, we then describe the mapping of the neural network to the hardware, \cref{subsec:nm_impl}.
Subsequently, we describe the \itl/ training in detail and demonstrate the application of this procedure to a handwritten digit recognition task, \cref{subsec:itl_training,sec:results}.

\section{The BrainScaleS Wafer-Scale System}
\label{sec:bss}
\begin{figure}[t]
  \captionsetup{farskip=0pt}
  \hfill
  \subfloat[\label{wafer_module:exploded_view}]{\def\svgwidth{.5\linewidth}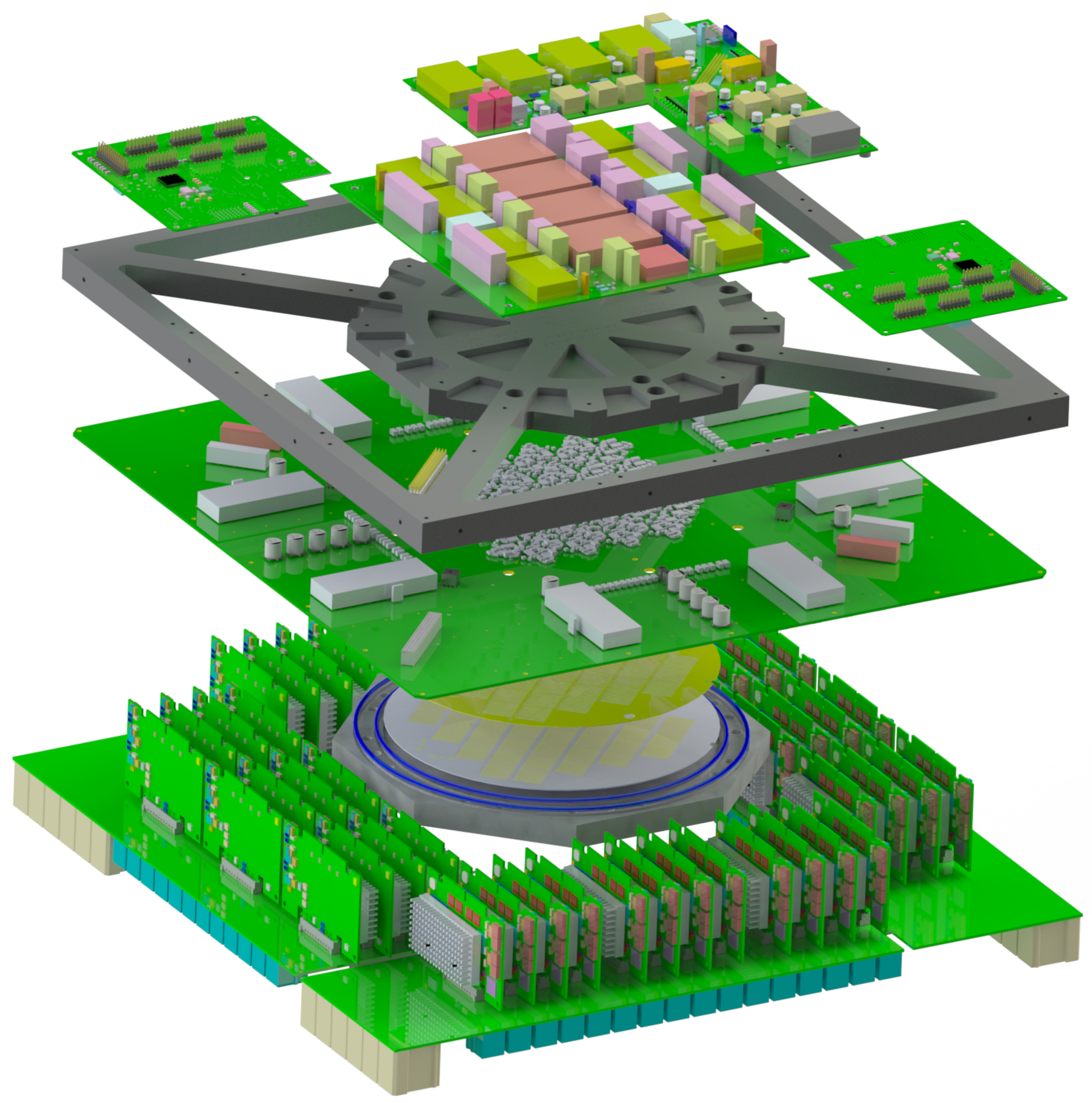}
  \hfill
  \subfloat[\label{wafer_module:picture}]{\includegraphics[height=0.2\textheight]{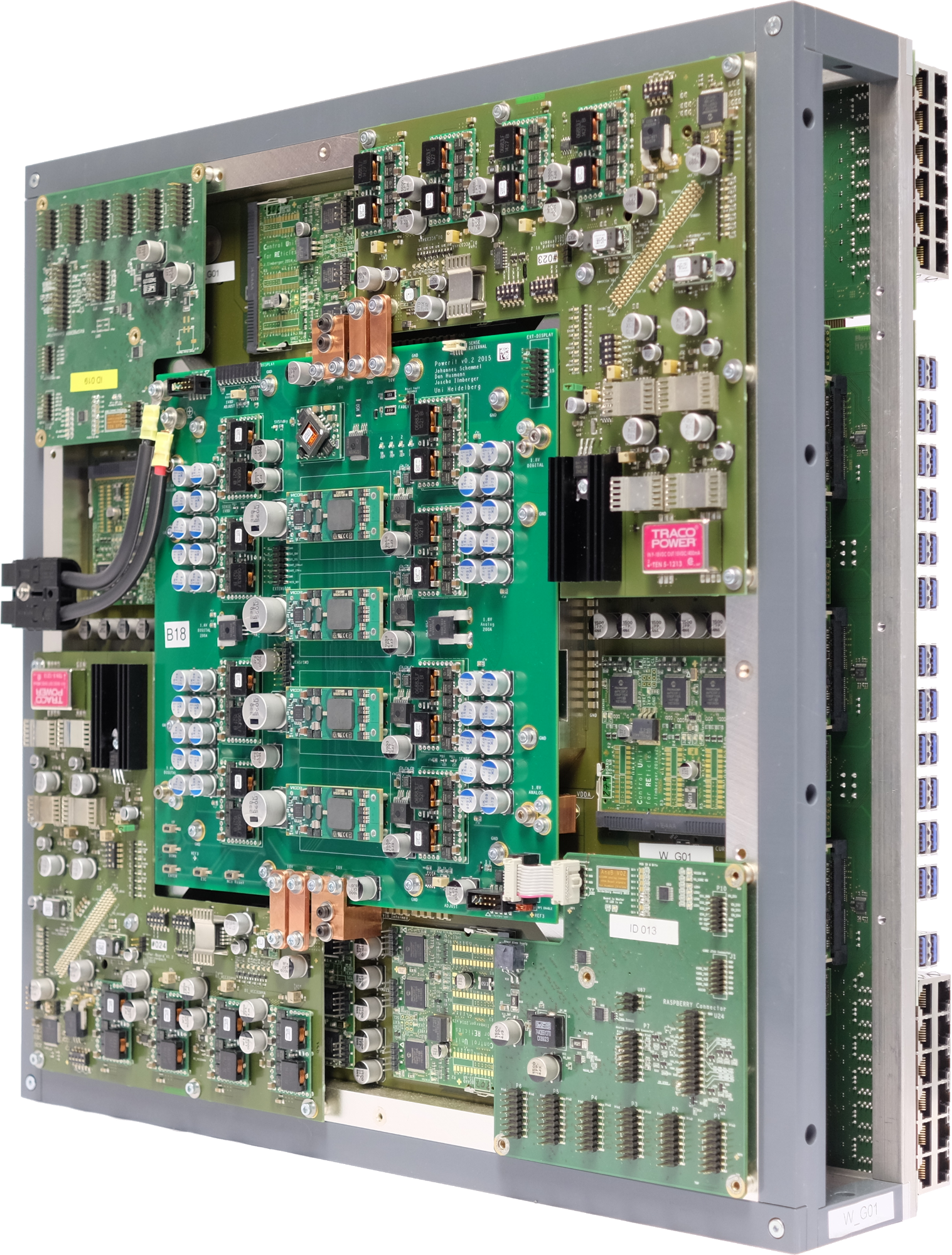}}
  \hspace*{\fill}
  \caption{%
    \protect\subref{wafer_module:exploded_view} 3D-schematic of a \BSSWM/ (dimensions: \SI{50}{\centi\meter} \texttimes{} \SI{50}{\centi\meter} \texttimes{} \SI{15}{\centi\meter}) hosting the wafer (A) and \num{48} FPGAs (B).
    The positioning mask (C) is used to align elastomeric connectors that link the wafer to the large main PCB (D).
    Support PCBs provide power supply (E \& F) for the on-wafer circuits as well as access (G) to analog dynamic variables such as neuron membrane voltages.
    The connectors for inter-wafer (USB slots) and off-wafer/host connectivity (Gigabit-Ethernet slots) are distributed over all four edges (H) of the main PCB.
    Mechanical stability is provided by an aluminum frame (I).
    \protect\subref{wafer_module:picture} Photograph of a fully assembled wafer module.
  }
  \label{fig:wafer_module}
\end{figure}

The \BSSS/ follows the principle of so-called ``physical modeling'', wherein the dynamics of VLSI circuits are designed to emulate the dynamics of their biological archetypes instead of numerically computing them as in the conventional simulation approach of von~Neumann architectures.
Neurons and synapses are implemented by analog circuits that operate in continuous time, governed by time constants which arise from the properties of the transistors and capacitors on the microelectronic substrate.
In contrast to real-time neuromorphic devices, see~\cite{liu2015event}, the analog circuits on our system are designed to operate in a regime where characteristic time constants (e.g., $\tausyn, \taum$) are much smaller than typical corresponding biological values.
This defines our intrinsic hardware acceleration factor of \num{10000} with respect to biological real-time.
The system is based on the ideas described in~\cite{schemmeliscas2010} but in the meantime it has advanced from a lab prototype to a larger installation comprising \num{20} wafer modules, see \cref{fig:machineroom}.

\subsection{The Wafer Module}

At the heart of the \BSSWM/, see \cref{fig:wafer_module}, is a silicon wafer with 384 HICANN (High Input Count Analog Neural Network) chips produced in \SI{180}{\nano\meter} CMOS technology.
It comprises 48 reticles, each containing 8 HICANNs, that are connected in a post-processing step.
Each chip hosts \num{512} neurons emulating Adaptive exponential integrate-and-fire (AdEx) dynamics~\cite{brette2005adex,millner2010AdEx} being able to reproduce most of the firing regimes discussed in~\cite{naud2008firing}.
When forming logical neurons by combining up to \num{64} neuron circuits, a maximum input from \num{14080} conductance-based synapses is reached where each circuit contributes with \num{220} synapses.

While the synapse and neuron dynamics are emulated by the analog circuits in continuous time,
action potentials are transported as digital data packets~\cite{schemmel_ijcnn2008}.
The action potentials, or spikes, are injected asynchronously into circuit-switched routing structures on the chip
and can be statically routed to target synapses and transported off-chip as time-stamped digital events via a packet-based network~\cite{thanasoulis2014pulse,scholze2011vlsi}.

\num{48} Xilinx Kintex-7 FPGAs, one per reticle, provide an I/O interface for configuration and spike data.
The connection between FPGAs and the control cluster network is established using standard Gigabit and 10-Gigabit-Ethernet.

Auxiliary PCBs provide the \BSSWSS/ with power, control and analog readout.

The specified maximum design power of a single module is \SI{2}{\kilo\watt}.
This operating point (MaxHW) assumes an average spike rate of \SI{40}{\hertz} applied to all hardware synapses.
As there are currently no power management techniques in use, all numbers reported in \cref{tab:hardware_utilization} are based on the maximum design power.
\Cref{tab:hardware_utilization} also provides data regarding hardware utilization for previously published neural network architectures~\cite{petrovici2014characterization}.

\begin{table}
  \centering
  \caption{Hardware utilization and power ratings for different neural network architectures.%
  }
  \label{tab:hardware_utilization}
  \begin{tabular}{@{}lrrr@{}}
    \toprule
    Model                 & L2/3 Model\textsuperscript{\cite{petrovici2014characterization}}
                          & AI Network\textsuperscript{\cite{petrovici2014characterization}}
                          & MaxHW                 \\
    \midrule
    HICANNs               & \num{352}            & \num{384}             & \num{384}             \\
    Neurons               & \num{14375}          & \num{22445}           & \num{196608}          \\
    Synapses              & \num{3470000}        & \num{4030000}         & \num{43253760}        \\
    Average Rate (Bio)    & \SI{4.8}{\hertz}     & \SI{13.6}{\hertz}     & \SI{40}{\hertz}       \\
    Speedup (Bio $\rightarrow$ HW)    & \num{12000}          & \num{10000}           & \num{10000}           \\
    Total Rate (HW)       & \SI{200}{\giga\hertz}& \SI{550}{\giga\hertz} & \SI{17.3}{\tera\hertz}\\
    Energy/Synaptic Event & \SI{10}{\nano\joule} & \SI{3.6}{\nano\joule} & \SI{0.1}{\nano\joule} \\
    \bottomrule
  \end{tabular}
\end{table}

\subsection{Running Neuronal Network Experiments}\label{subsec:software_stack}

The \BSS/ software stack transforms a user-defined abstract neural network description, i.e., network topology, model parameters and input stimuli, to a corresponding hardware-constrained experiment configuration.

Descriptions of spiking neural networks are often formulated using dedicated languages.
Most are based on either declarative syntax, e.g., NineML~\cite{raikov2011nineml} or NeuroML~\cite{gleeson2010neuroml}, or use procedural syntax, e.g., the Python-based API called PyNN~\cite{davison08pynn}.
The current \BSSS/ uses PyNN to describe neural network experiments based on experiences with previous implementations~\cite{bruederle09establishing}.
This design choice enables the use of the versatile software packages developed in the PyNN ecosystem, such as the Connection Set Algebra~\cite{djurfeldt2012csa}, Elephant~\cite{denker2015elephant} or Neo~\cite{garcia2014neo}.

\begin{figure}[t]
  \includegraphics[width=0.49\linewidth]{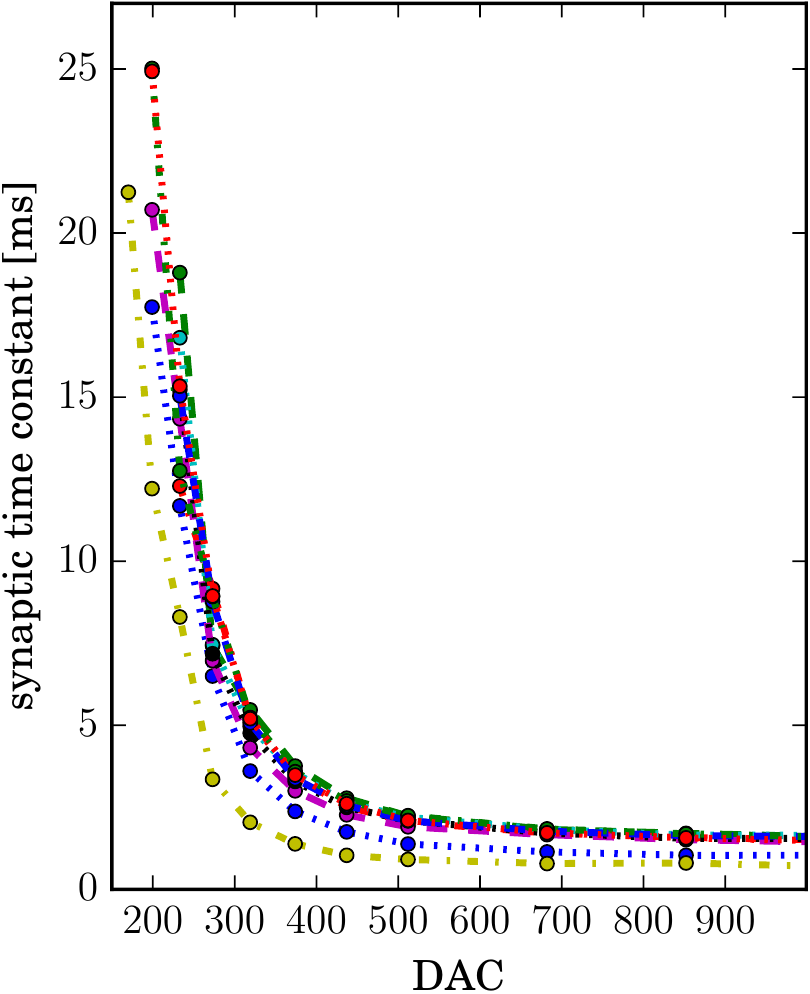}\hfill%
  \includegraphics[width=0.49\linewidth]{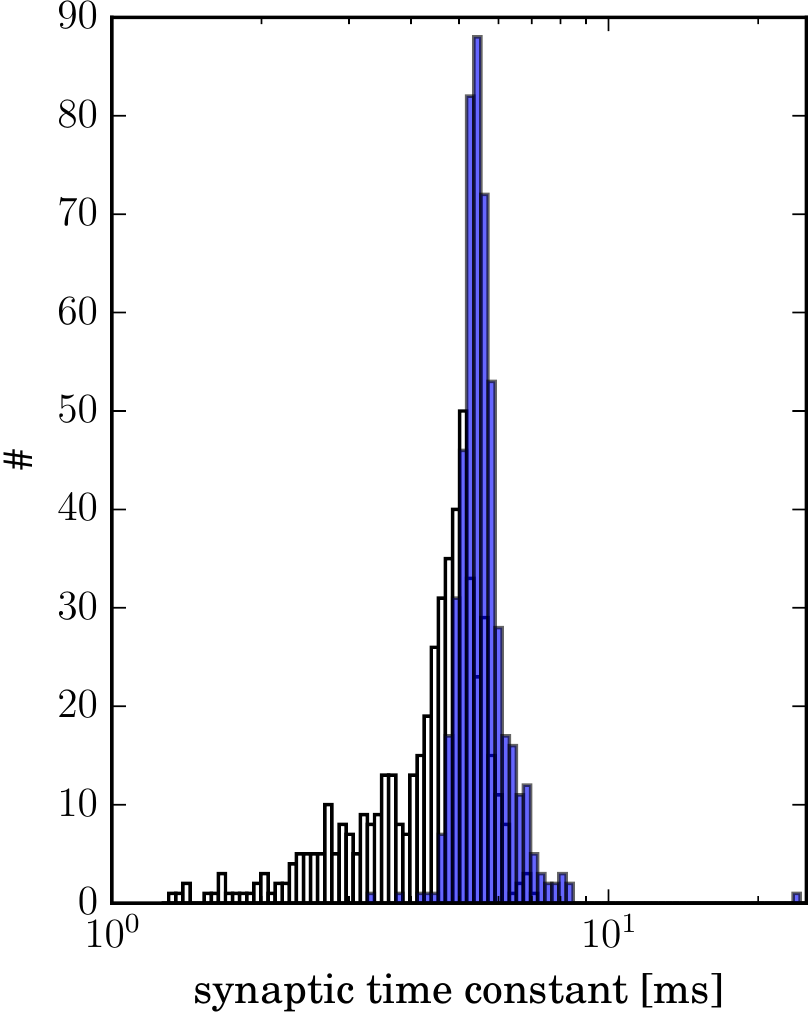}
  \caption{Example for the calibration of the synaptic time constant. Left: measured synaptic time constants (y-axis) for different neurons as a function of the digital parameter (DAC, x-axis) controlling the responsible analog parameter.
    Right: measured synaptic time constant with (blue) and without (white) calibration for all neurons of a HICANN (right).
  }
  \label{fig:calib_tausyn}
\end{figure}

Starting from the user-defined experiment description in PyNN, the transformation process maps model neurons to hardware circuits, routes connections between neurons to create synapses, and translates the model parameters to hardware settings.
This translation of neuron and synapse model parameters requires calibration data, see \cref{subsec:calibration}, as well as rules for the conversion between the biological and the hardware time and voltage domains.

The result of the whole transformation process is a hardware-compatible, abstract experiment description which can be converted into low-level configuration data.
After acquiring hardware access using a fair resource scheduling and queuing system based on SLURM~\cite{jette2003slurm},
the hardware is configured and the experiment is ready to run on the system.

Although the \BSS/ software stack provides a user-friendly modeling interface and hides hardware specifics, all low-level settings are available to the expert user.
In particular the experiments presented here make use of this feature, enabling fast iterative modification of synaptic weights and input stimuli.

\subsection{Calibration}\label{subsec:calibration}

\begin{figure}[t]
  \centering
  \includegraphics[width=0.95\linewidth]{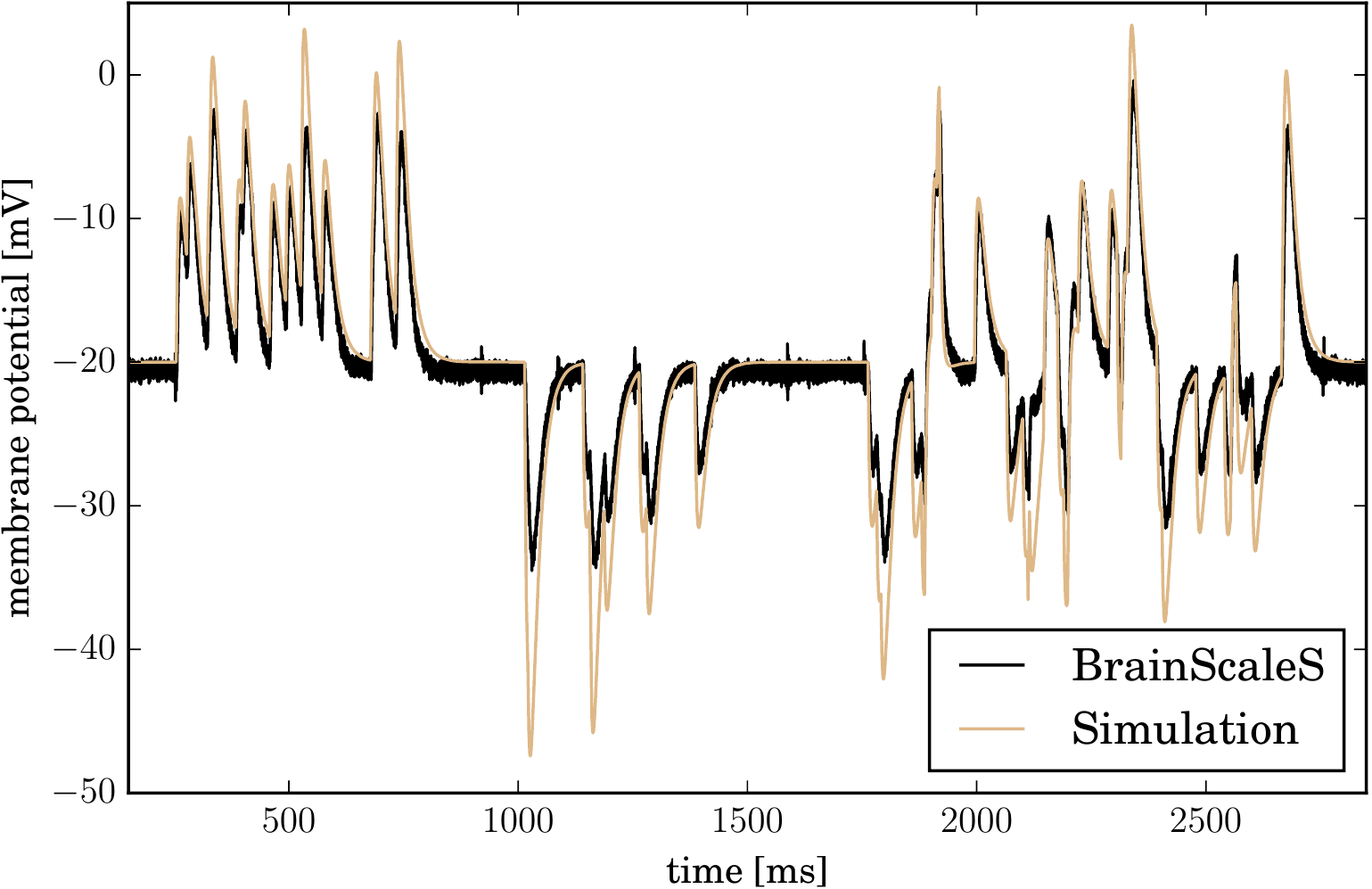}
  \caption{Comparison of a recorded membrane trace to a neuron simulated with NEST.
           The neuron receives an excitatory Poisson stimulus of \SI{20}{\hertz} followed by inhibitory and then simultaneous excitatory and inhibitory Poisson stimuli of the same frequency.
           All calibrations are applied and the hardware response is converted to the emulated biological domains.%
         }
  \label{fig:calib_trace}
\end{figure}

For each neuron, the calibration provides translation rules from target parameters, such as the membrane time constant, to a set of corresponding hardware control parameters.
Thereby it accounts for circuit-to-circuit variations caused by the transistor mismatch inherent to the wafer manufacturing process.
The data are stored in the hardware domains and two scaling rules are used for the conversion to the biological time and voltage domains.
All time constants are scaled with the acceleration factor of \num{10000}, e.g., \SI{1}{\micro\second} hardware time corresponds to \SI{10}{\milli\second} of emulated biological time.
Voltages are scaled according to
\begin{equation}
  \label{eq:bio_to_hw_potentials}
  V_{\text{hardware}} = V_{\text{bio}} \times \alpha + s,
\end{equation}
where $\alpha$ is a unit-free scaling factor and $s$ is an offset.
From here on, all units are given in the biological domain if not stated otherwise.

\Cref{fig:calib_tausyn} exemplifies the calibration technique for the particular case of the synaptic time constant.
For every neuron, the analog parameter controlling the synaptic time constant is varied and the resulting synaptic time constant is determined from a recorded post-synaptic potential.
A fit to this data then provides the mapping from the desired synaptic time constant to the value of the control parameter.
Calibration reduces the neuron-to-neuron variation significantly, but not perfectly.
The remaining variability is mostly caused by the trial-to-trial variation of the analog parameter storage.

\Cref{fig:calib_trace} shows two membrane time courses comparing a calibrated silicon neuron to a numerical simulation with NEST~\cite{Gewaltig:NEST}.
In both cases, the same model parameters and input spike trains were used.
Despite the overall match, it can be seen that the calibration is not perfect, e.g.\, for the neuron used in \cref{fig:calib_trace}, the inhibitory stimulus is weaker compared to the expectation from simulation.
Due to the analog nature of the system, variations will always occur to a certain extent, rendering \itl/ training essential for networks that are sensitive to parameter noise, as we discuss in the following.

\section{Training a Deep Spiking Network}

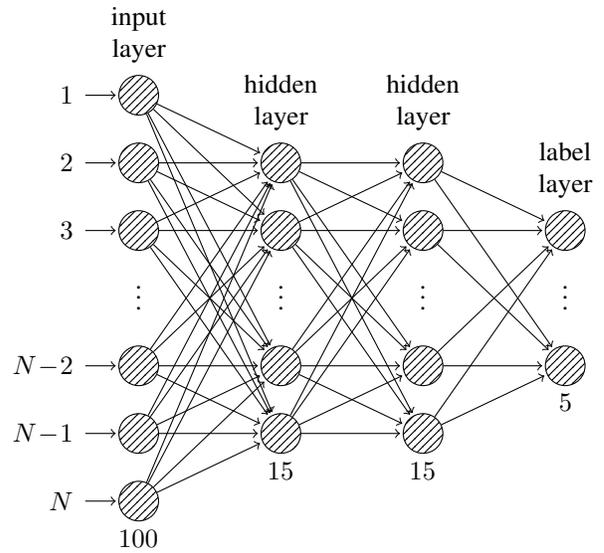
\begin{figure}[t]
  \centering

  \def\layersephoriz{2.10cm}
  \def\layersepvert{0.5cm}

  \begin{tikzpicture}[shorten >=1pt,->,draw=black, node distance=\layersephoriz, scale=0.9]
    \tikzstyle{every pin edge}=[<-,shorten <=1pt]
    \tikzstyle{neuron}=[circle,draw,fill=black!25,minimum size=15pt,inner sep=0pt]
    \tikzstyle{input neuron}=[neuron, pattern=north east lines];
    \tikzstyle{hidden neuron}=[neuron, pattern=north east lines];
    \tikzstyle{output neuron}=[neuron, pattern=north east lines];
    \tikzstyle{annot} = [text width=4em, text centered]

    \foreach \name / \y in {$1$/1,$2$/2,$3$/3,$N\!-\!2$/5,$N\!-\!1$/6,$N$/7}
    \path[yshift=3*\layersepvert]
    node[input neuron, pin=left:\name] (I-\y) at (0,-\y) {};

    \path[]
    node[] (I-4) at (0, -5*\layersepvert+\layersepvert/4) {\vdots};

    \foreach \layer / \x in {1,2}
    \foreach \name / \y in {1,2,4,5}
    \path[yshift=\layersepvert]
    node[hidden neuron] (H\layer-\name) at (\layer*\layersephoriz,-\y cm) {};

    \foreach \layer / \x in {1,2}
    \path[]
    node[] (H\layer-3) at (\layer*\layersephoriz, -5*\layersepvert+\layersepvert/4) {\vdots};

    \foreach \name / \y in {1,3}
    \path[yshift=-\layersepvert]
    node[output neuron] (O-\name) at (3*\layersephoriz,-\y cm) {};

    \path[]
    node[anchor=center] (I-4) at (3*\layersephoriz, -5*\layersepvert+\layersepvert/4) {\vdots};


    \foreach \source in {1,2,3,5,6,7}
    \foreach \dest in {1,2,4,5}
    \path (I-\source) edge (H1-\dest);

    \foreach \source in {1,2,4,5}
    \foreach \dest in {1,2,4,5}
    \path (H1-\source) edge (H2-\dest);

    \foreach \source in {1,2,4,5}
    \foreach \dest in {1,3}
    \path (H2-\source) edge (O-\dest);


    \node[annot,above of=I-1, node distance=0.8cm] (il) {input layer};
    \node[annot,above of=H1-1, node distance=0.8cm] (hl1) {hidden layer};
    \node[annot,above of=H2-1, node distance=0.8cm] (hl2) {hidden layer};
    \node[annot,above of=O-1, node distance=0.8cm] {label layer};

    \node[annot,below of=I-6, yshift=0.7cm]   {$100$};
    \node[annot,below of=H1-4, yshift=0.7cm]   {$15$};
    \node[annot,below of=H2-4, yshift=0.7cm]   {$15$};
    \node[annot,below of=O-3,  yshift=2.3*0.7cm] {$5$};

  \end{tikzpicture}
  \caption{Topology of the feed-forward neural network with one input layer, two hidden layers and one label layer.
    The dimension of the input layer is equal to the number of pixels of the input image.
    The number of label units is equal to the number of image classes the network is trained to recognize.
  }
  \label{fig:dnn}
\end{figure}

In the following, we describe our network model and training setup.
Since we are using an abstract network of rectified linear units (ReLUs) and an equivalent spiking network of leaky integrate-and-fire (LIF) neurons in parallel, we will first describe the networks structure in abstract terms.

Our network is modeled as a feed-forward directed graph as shown in \cref{fig:dnn}.
The input layer, consisting of \num{100} units, is used to represent the input patterns that the network later learns to classify.
Each of these classes is represented by one label unit.
Between the input and label layers are two 15-unit hidden layers that learn particular features in the input space.
The weights of the directed edges are learned during several phases of training, as described farther below.

Our network was trained on a modified subset of the MNIST dataset of handwritten digits~\cite{lecunmnist}.
First, we decreased the resolution from \num{28 x 28} pixels to \num{10 x 10} pixels by bicubic interpolation.
To account for the lower dissimilarity of the reduced resolution images, we restricted the dataset to the five digit classes \class{0}, \class{1}, \class{4}, \class{6} and \class{7}.
This results in a training set of \num{30690} and a test set of \num{5083} images.

The spiking neural network is then trained in three phases:

\begin{enumerate}
    \renewcommand{\theenumi}{\Alph{enumi}}
    \renewcommand\labelenumi{\theenumi.}
    \item The software model of the network with rectified linear units (ReLUs) is trained with classical backpropagation.
    \item The resulting weights are converted to synaptic weights in an appropriately parametrized LIF network on the BrainScaleS hardware.
    \item The synaptic weights are further trained in a hardware-software training loop.
\end{enumerate}

\begin{figure}[t]
  \centering
  \includegraphics[width=\linewidth]{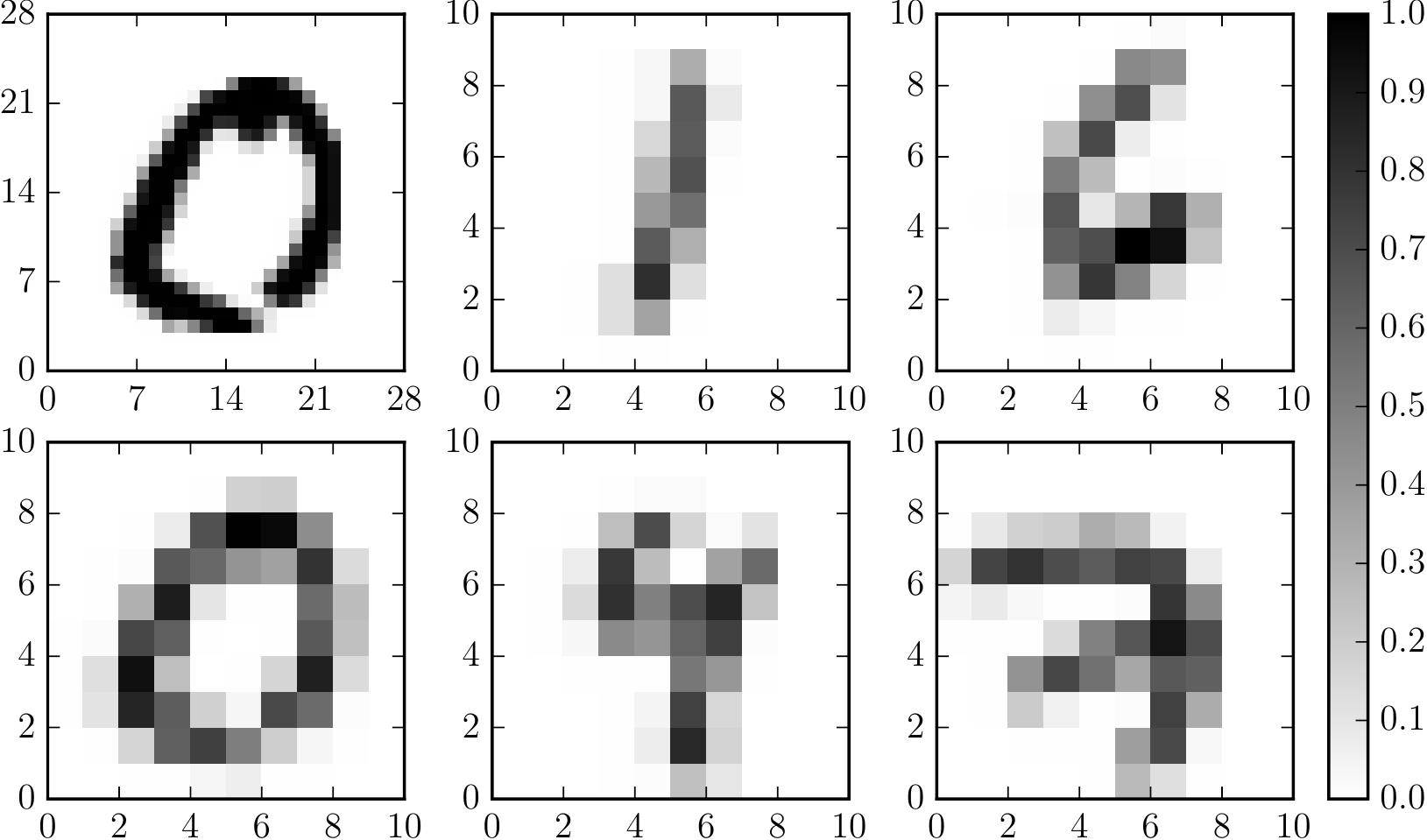}
  \caption{Examples of the input data used during training.
    Original MNIST image of a \class{0} (upper left) vs.\ reduced-resolution image (lower left).
    Middle and right column: reduced-resolution images from the other four classes (\class{1}, \class{4}, \class{6}, \class{7}).
    }
  \label{fig:reduced_mnist}
\end{figure}

\subsection{Software Model}\label{subsec:initialtraining}

The training of the software model is performed similarly to~\cite{Cao2015} using the TensorFlow~\cite{tensorflow2015-whitepaper} software with the properties detailed in the following.

\subsubsection{Input}
The grayscale value of the input image pixels is transformed to a number between 0 and 1 and set as the activation of the units in the input layer.

\subsubsection{Units}
The output $x_k$ of ReLU unit $k$ is given by
\begin{equation}
  x_k = R\Big(\sum_l W_{kl} x_l\Big),\quad R(a) = \max(0,a), \label{eq:relu}
\end{equation}
where $W_{kl}$ is the weight of the connection from unit $l$ to unit $k$,
$R: \mathbb{R}\rightarrow\mathbb{R}$ is the activation function of a ReLU, and the sum runs over all indices $l$ of units from the previous layer.

\subsubsection{Weights}
The initial weights for layer $n$ containing $N_n$ units are drawn from a normal distribution with a mean of zero and standard deviation
\begin{equation}
  \label{eq:truncated_normal}
  \sigma_n = \frac{1}{\sqrt{N_{n-1}}},
\end{equation}
where weight magnitudes $> 2\sigma_n$ are dropped and re-picked.

\subsubsection{Training}
The network is trained by mini-batch gradient-descent with momentum~\cite{Qian1999145} minimizing the cost function
\begin{equation}
  \label{eq:cost}
  C(W) = \tfrac{1}{5} \sum_{s \in S} {(\mathbf{\tilde{y}}_{s} - \mathbf{\hat{y}}_{s})}^2 + \sum_{kl} \tfrac{1}{2}\lambda W_{kl}^2,
\end{equation}
where $W$ is the matrix containing all network weights,
$\mathbf{\hat{y}}_{s}$ the one-hot vector for the true digit, $\mathbf{\tilde{y}} = \frac{\mathbf{y}}{N_{\text{2nd hidden}}} $ the scaled activity of the label layer and $S$ the samples in the current batch of \num{100} samples.

The first term in \cref{eq:cost} is the euclidean distance
between the predicted labels $\mathbf{\tilde{y}}$ and the true labels $\mathbf{\hat{y}}$, rewarding correct and penalizing wrong activity.
The second term of \cref{eq:cost} regularizes the weights with $\lambda=0.001$, leading to the suppression of large weights to prevent overfitting.

Per training step, the weights are updated according to
\begin{align}
  \label{eq:gradientdescent}
  \Delta W_{kl} &\leftarrow \eta \nabla_{W_{kl}} C(W) + \gamma \Delta W_{kl},\\
  W_{kl} &\leftarrow W_{kl} - \Delta W_{kl},
\end{align}
where $\Delta W_{kl}$ is the change in weight, $\eta=0.05$ is learning rate, and $\gamma=0.9$ the momentum parameter.
In foresight of the hardware implementation, $W_{kl}$ is clipped to~$[-1,1]$.

\subsection{Neuromorphic Implementation}\label{subsec:nm_impl}

\begin{table}
  \centering
  \caption{Neuron parameters and typical post-calibration variations.%
  }
  \label{tab:neuron_parameters}
  \begin{tabular}{@{}rrc@{}}
    \toprule
    Parameter                         & Value                    & Relative Variation\\\midrule
    Inhibitory Reversal Potential     & \SI{-80}{\milli\volt}    & \SI{5}{\percent}  \\
    Reset Potential                   & \SI{-64}{\milli\volt}    & \SI{2}{\percent}  \\
    Resting Potential                 & \SI{-40}{\milli\volt}    & \SI{10}{\percent} \\
    Spike Threshold                   & \SI{-37.5}{\milli\volt}  & \SI{0.5}{\percent}\\
    Excitatory Reversal Potential     & \SI{0}{\milli\volt}      & \SI{0.5}{\percent}\\
    Inh./Exc.\ Synaptic Time Constant & \SI{5}{\milli\second}    & \SI{10}{\percent} \\
    Membrane Time Constant            & \SI{20}{\milli\second}   & \SI{10}{\percent} \\
    \bottomrule
  \end{tabular}
\end{table}

\subsubsection{Input}

The input image is converted to Poisson spike trains following~\cite{10.3389/fnins.2013.00178}:
\begin{equation}
\label{eq:inputrates}
\nu_p = \frac{c_p}{\sum_p c_p} \cdot \nu_\tot,
\end{equation}
where $\nu_p$ is the firing rate of the input corresponding to the $p$th pixel, $c_p$ the grayscale value of the $p$th pixel and $\nu_\tot$ is the targeted total firing rate the input layer receives.
In our case, we set $\nu_\tot = \SI{2500}{Hz}$.
Each pattern is presented for \SI{0.9}{\second} followed by \SI{0.1}{\second} of silence to allow the activity to decay.

\subsubsection{Hardware Configuration}

The network is mapped to the \BSS/ hardware using the software stack detailed in~\cref{subsec:software_stack}.
Neurons of all layers, including the input layer, are randomly placed on \num{8}~HICANNs.
For input and on-wafer routing, \num{6} additional chips are used.
These \num{14} HICANNs are connected to \num{5} different FPGAs.
For each artificial neuron, four hardware neuron circuits are connected to form one logical neuron to increase the number of possible inputs.
Except for the stimulus to the input layer, each pair of neurons in consecutive layers is connected with both an inhibitory and an excitatory synapse.
This allows the weights to change sign during learning without having to change the configured topology.
Therefore, the hardware-emulated network has a total of 3700 synapses.

\subsubsection{Neuron parameters}

Despite the different input and output domain, the activation functions of ReLUs and LIF neurons share features, i.e., both have a threshold below which the output is zero and a positive gradient for suprathreshold input.
Not all neuron features are required to mimic the ReLU behavior, therefore we disable the adaptation and exponential features of the AdEx model.
The parameters and the neuron-to-neuron variation after calibration, see~\cref{subsec:calibration}, are listed in \cref{tab:neuron_parameters}.
To allow a balanced representation of positive and negative weights, the reversal potentials have been chosen as symmetric around the resting potential.
The refractory period is set as small as possible to be close to a linear relation of the input to the output activity.
\Cref{eq:bio_to_hw_potentials} is used to convert to hardware units with $\alpha=20$ and $s = \SI{1800}{\milli\volt}$, e.g., the target value of the resting potential on hardware equals \SI{1}{\volt}.

\subsubsection{Weights}
The trained weights $W_{kl}$ of the \TFnet/ are converted to the \SI{4}{\bit} hardware weights $W'_{kl}$ by
\begin{equation}
  \label{eq:weight_scaling}
  W'_{kl} = \text{round}(|W_{kl}| \times 15).
\end{equation}
Positive (negative) weights are assigned to excitatory (inhibitory) synapses and the corresponding inhibitory (excitatory) synapse is turned off.

\subsection{Hardware \ITLnodashes/}\label{subsec:itl_training}

\begin{figure}[t]
  \centering
    \begin{tikzpicture}[
            ->,
            scale=1,
            line width=2pt,
            >=latex,
            transform shape,
        ]
        \def \radius {2.5cm}
        \pgfmathsetmacro{\aangle}{acos(1.0/3.0)} 
        \def \angleoffset {90} 
        \def\data{{{"backpropagation",0,28,8},{"weight updates",\aangle,8,8},{"4\,bit weight discretization",180-\aangle,8,22},{"BrainScaleS",180,22,8},{"spikes",180+\aangle,8,8},{"ReLU activity",360-\aangle,10,28}}}
        \def\numdata {6}

        \foreach \s [count=\i from 0] in {1,...,\numdata}
        {
            \pgfmathsetmacro{\c}{\data[\i][1] + \angleoffset}
            \node[fill=white,inner sep=1pt] (\s) at (\c:\radius) {\pgfmathparse{\data[\i][0]}\pgfmathresult};
            \pgfmathsetmacro{\arcstart}{mod(\c+\data[\i][2],360)}
            \pgfmathsetmacro{\arcend}{\data[mod(\s,\numdata)][1]+\angleoffset-\data[\i][3]}
            \draw[->] (\arcstart:\radius) arc (\arcstart:\arcend:\radius);
        }
        \node[anchor=base] (7) [left=of 4]  {MNIST};
        \node[anchor=base] (8) [right=of 4]  {prediction};
        \draw (7) -- (4);
        \draw (4) -- (8);
        \node (9) [below=-2pt of 7] {};
        \draw[color=gray!50] (9) --  (9-|8.south);
        \node[below=.1cm of 4,color=gray!70] {forward pass};
        \begin{scope}[on background layer]
            \pgfmathsetmacro{\arcstart}{-35}
            \pgfmathsetmacro{\arcend}{215}
            \def\radiusmult {1.14}
            \draw[line width=2pt,color=gray!50] (\arcstart:\radiusmult*\radius) arc (\arcstart:\arcend:\radiusmult*\radius);
            \node[above=3pt of 1,color=gray!70] {backward pass};
        \end{scope}
    \end{tikzpicture}
  \caption{
        Illustration of our \itl/ training procedure.
        In an antecedent step (not shown), a software-trained ReLU network, see \cref{fig:dnn}, is mapped to an equivalent LIF network on the \BSS/ hardware.
        Each iteration of in-the-loop training consists of two passes.
        In the forward pass, the output firing rates of the LIF network are measured in hardware.
        In the backward pass, these rates are used to update the synaptic weights of the LIF network by computing the corresponding weight updates in the ReLU network and mapping them back to the hardware.
        }
  \label{fig:intheloop}
\end{figure}
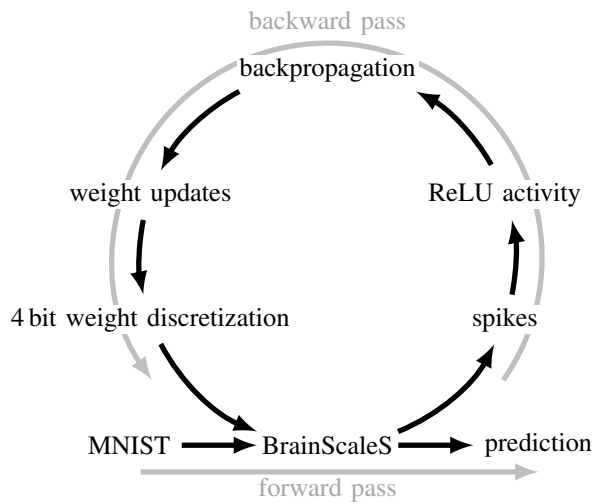

\Cref{subsec:nm_impl} laid out the necessary steps to convert the \TFnet/ to a network of neurons in analog hardware.
After conversion it was found that the classification accuracy was significantly reduced compared to the initially trained ANN.
To compensate for the reduced classification accuracy, training was continued with the hardware \itlnodashes/, see \cref{fig:intheloop}.
\Itl/ training consists of a series of training steps, each of which is performed as follows.
First, the neuron activity is recorded for a batch of training samples.
These firing rates are then equated to the ReLU unit response, where we used the following heuristic for the label layer: $\tilde{\mathbf{y}} = \frac{\mathbf{y}}{\SI{30}{\hertz}}.$
The resulting vector is used to compute the cost function $C(W)$ defined in \cref{eq:cost}.
The weight updates are then computed using \cref{eq:gradientdescent} using the ReLU activation function \cref{eq:relu} as an approximation of the difficult-to-determine activation functions of the hardware-emulated neurons.
For the experiments described here, we used the parameters $\eta=0.05$, $\gamma=0$ and a batch size of \num{1200} samples.

\section{Results}
\label{sec:results}

\begin{figure*}
  \centering
  \includegraphics[width=\linewidth]{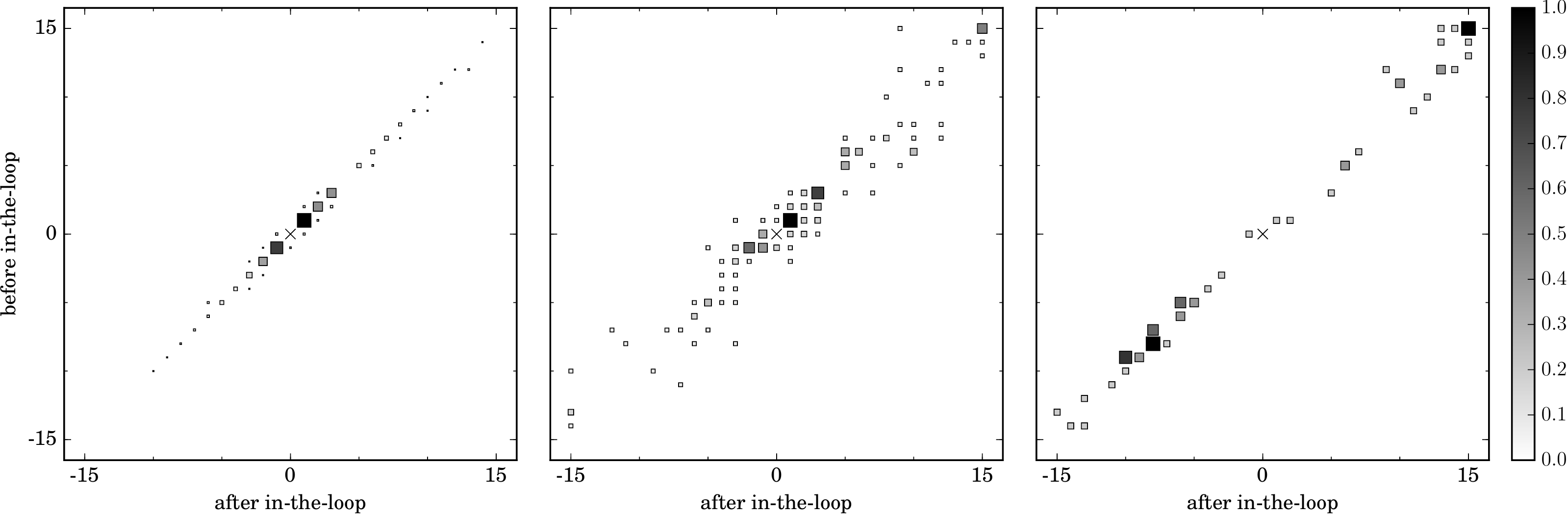}
  \caption{Correlation of hardware weights before and after \itl/ training for the projections to the first (left) and second hidden layer (center) and the label layer (right).
    Weights that are zero before and after training are omitted.
    The relative frequency is encoded by both grayscale and area of the corresponding square.
  }
  \label{fig:mnist_weight_migration}
\end{figure*}

\begin{figure*}[t]
  \centering
  \includegraphics[width=\linewidth]{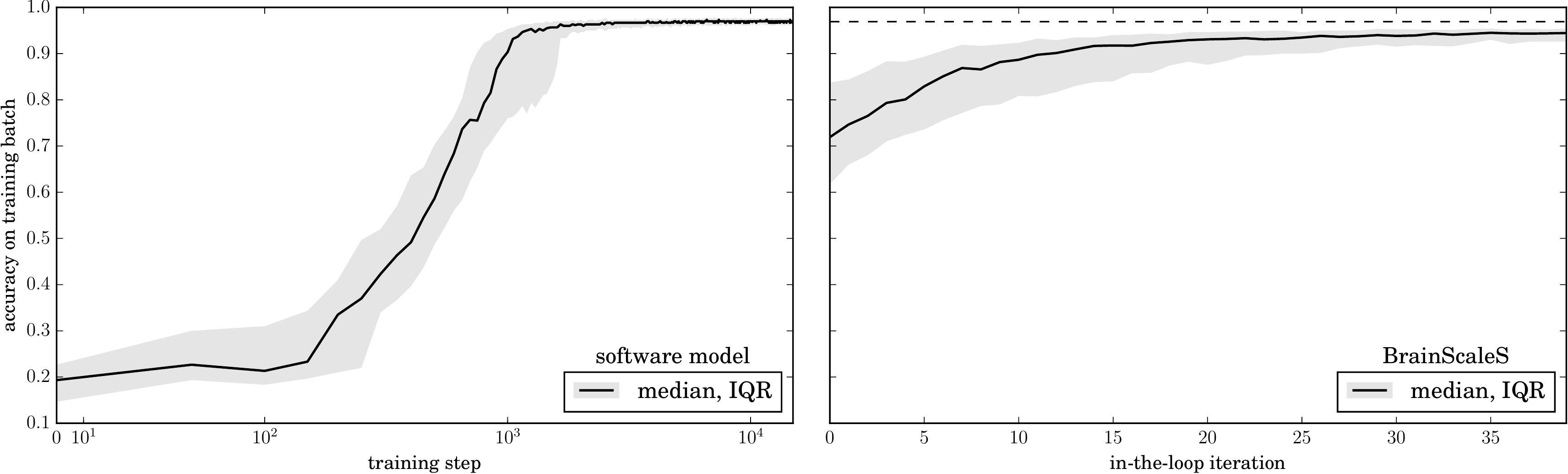}
  \caption{Classification accuracy per batch as a function of the training step for the software model (left) and the \itl/ iteration for the hardware implementation (right) for \num{130} runs.
    The uncertainty, given by the interquartile range (IQR), expresses the variations when repeating the software model with different initial weights and the \itl/ training using different initial weights for the ReLU training and different sets of hardware neurons.
  }
  \label{fig:mnist_accuracy}
\end{figure*}

\begin{figure*}[t]
  \centering
    \begin{overpic}[height=\textwidth, angle=-90]{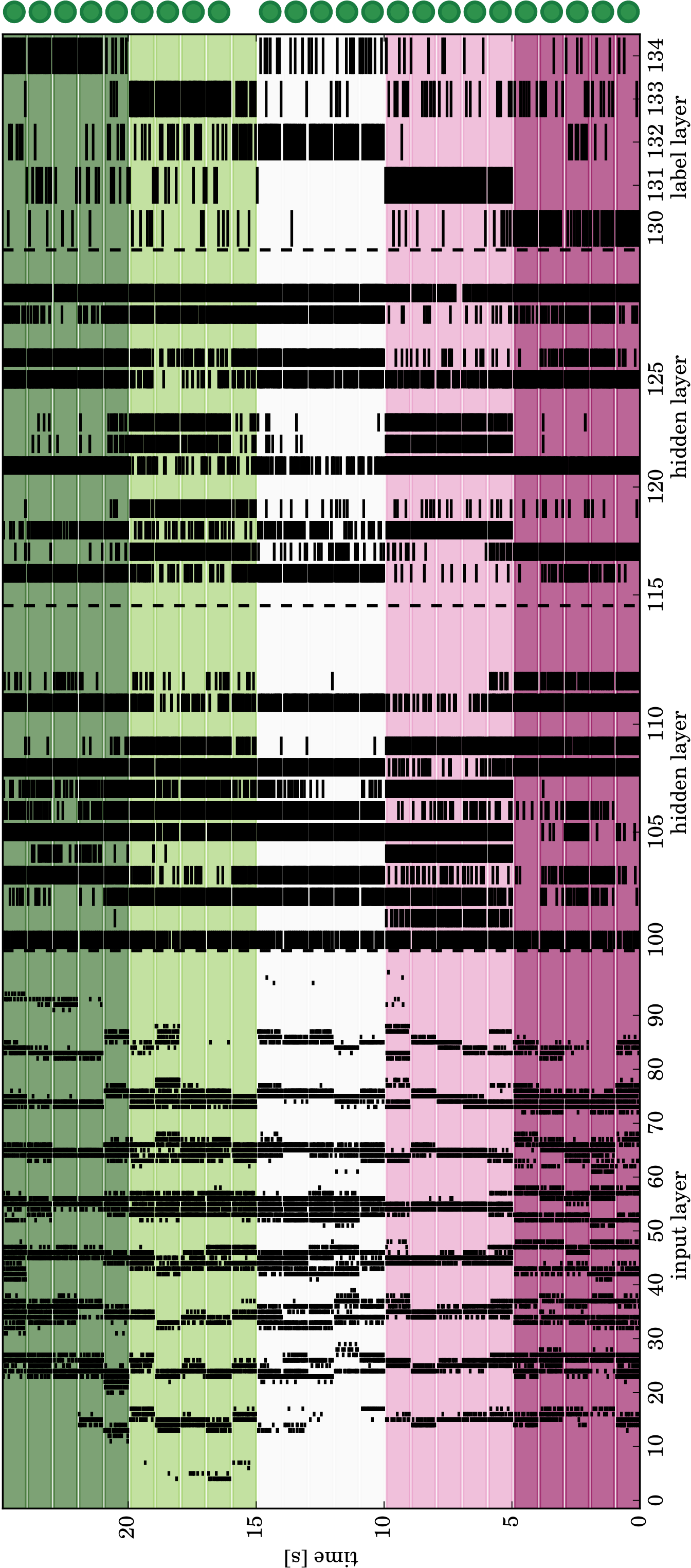}
      \put(15,48){\includegraphics[width=0.9cm]{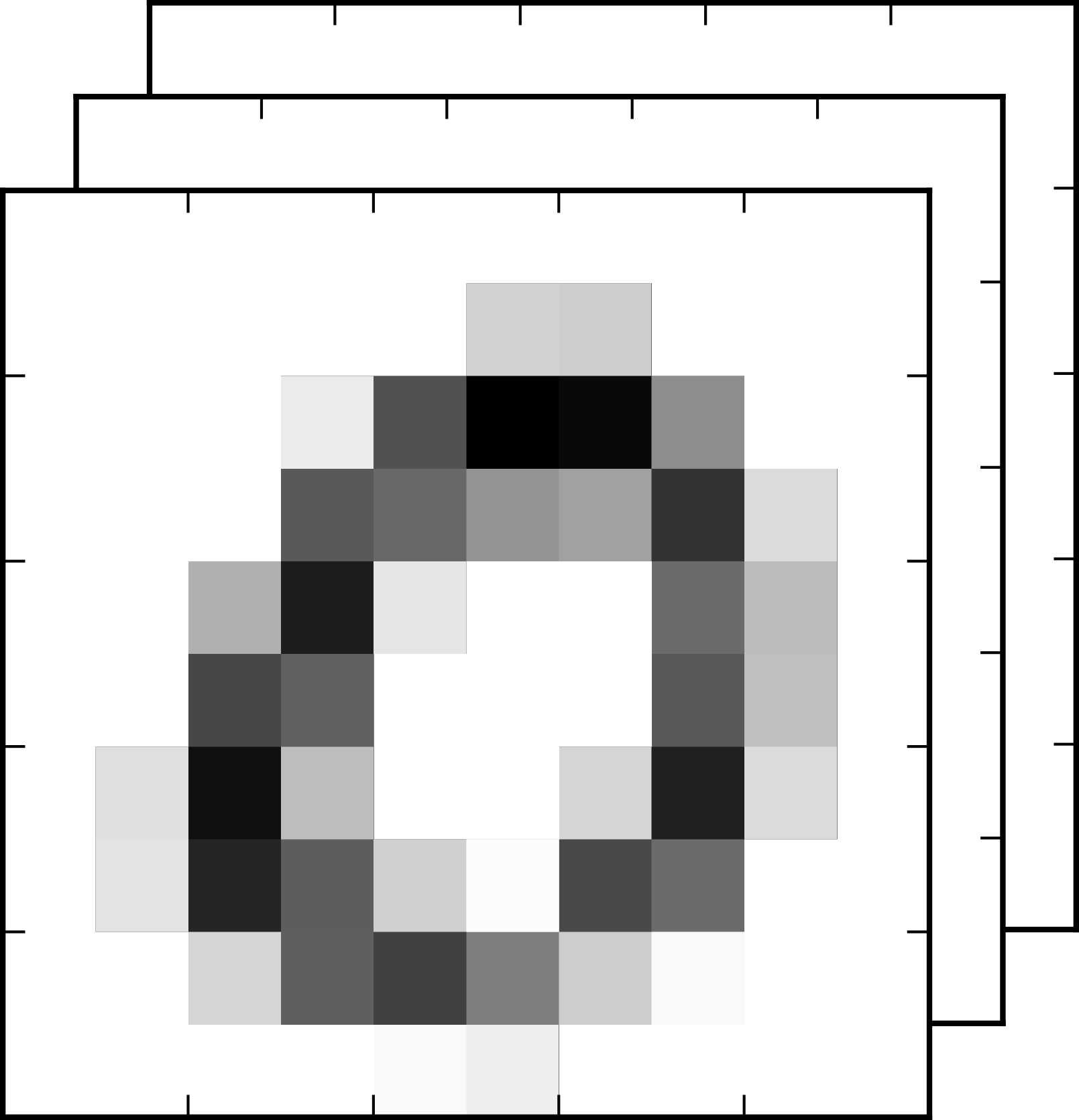}}
      \put(15,130){\includegraphics[width=0.9cm]{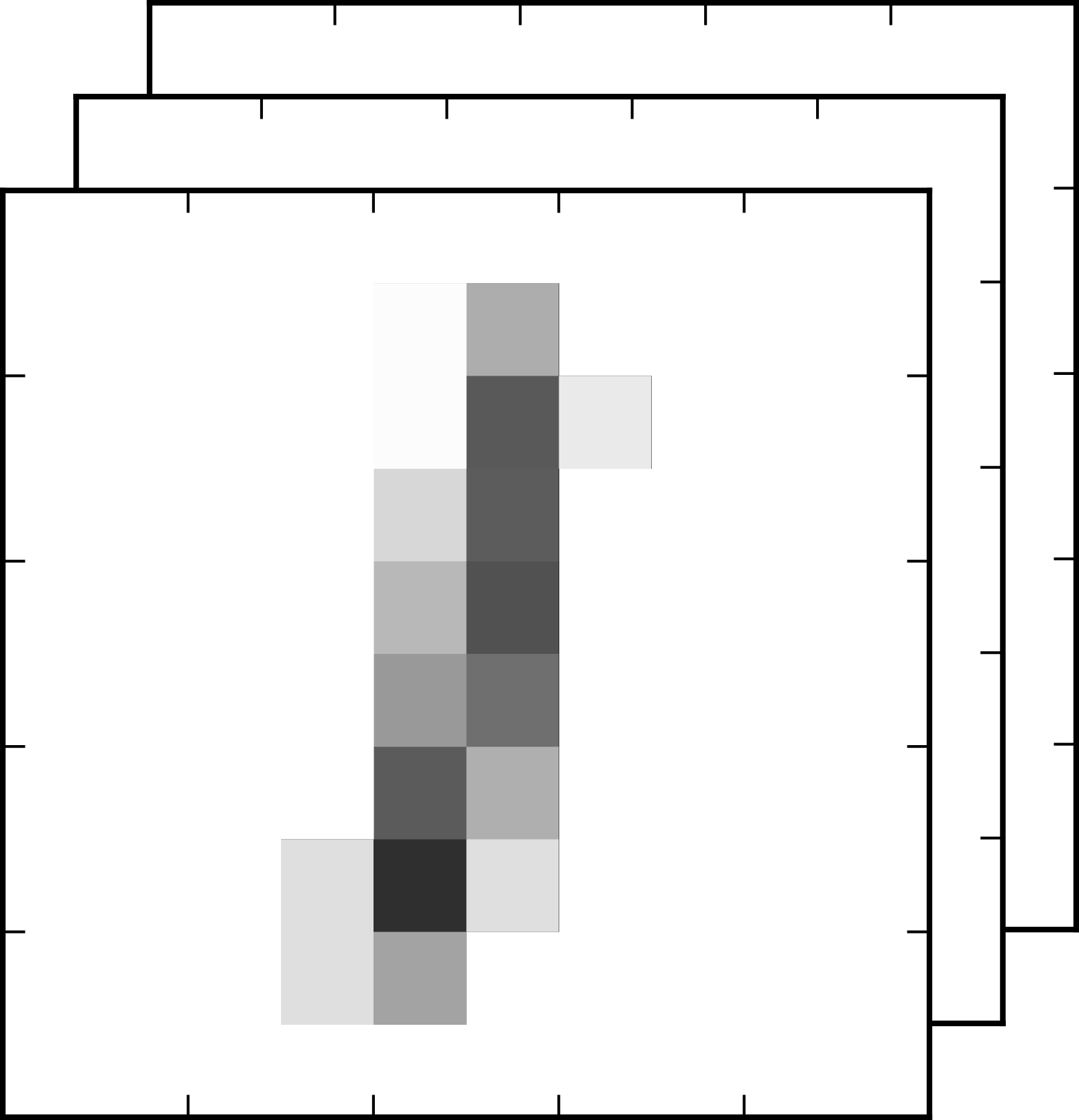}}
      \put(15,212){\includegraphics[width=0.9cm]{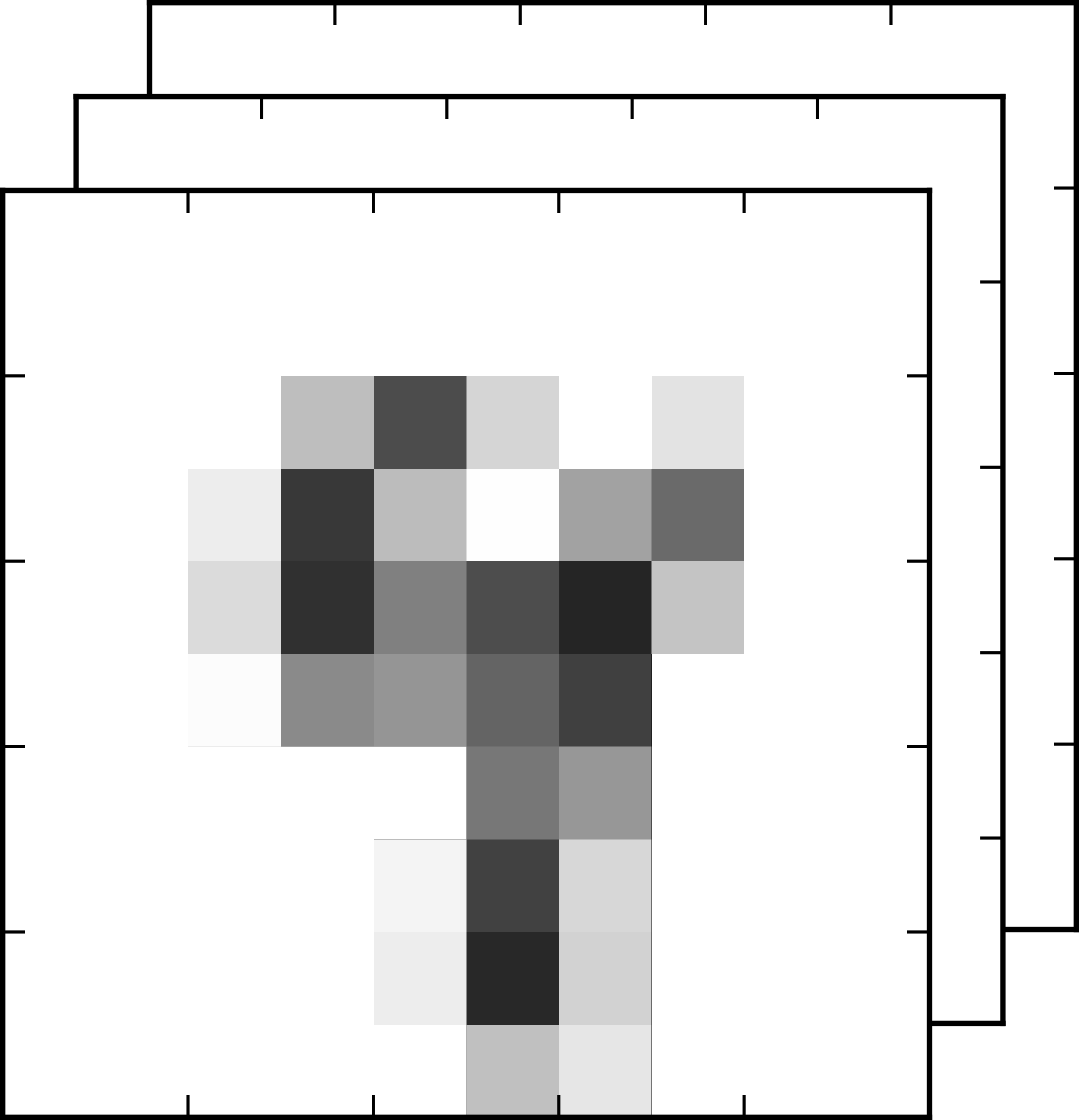}}
      \put(15,294){\includegraphics[width=0.9cm]{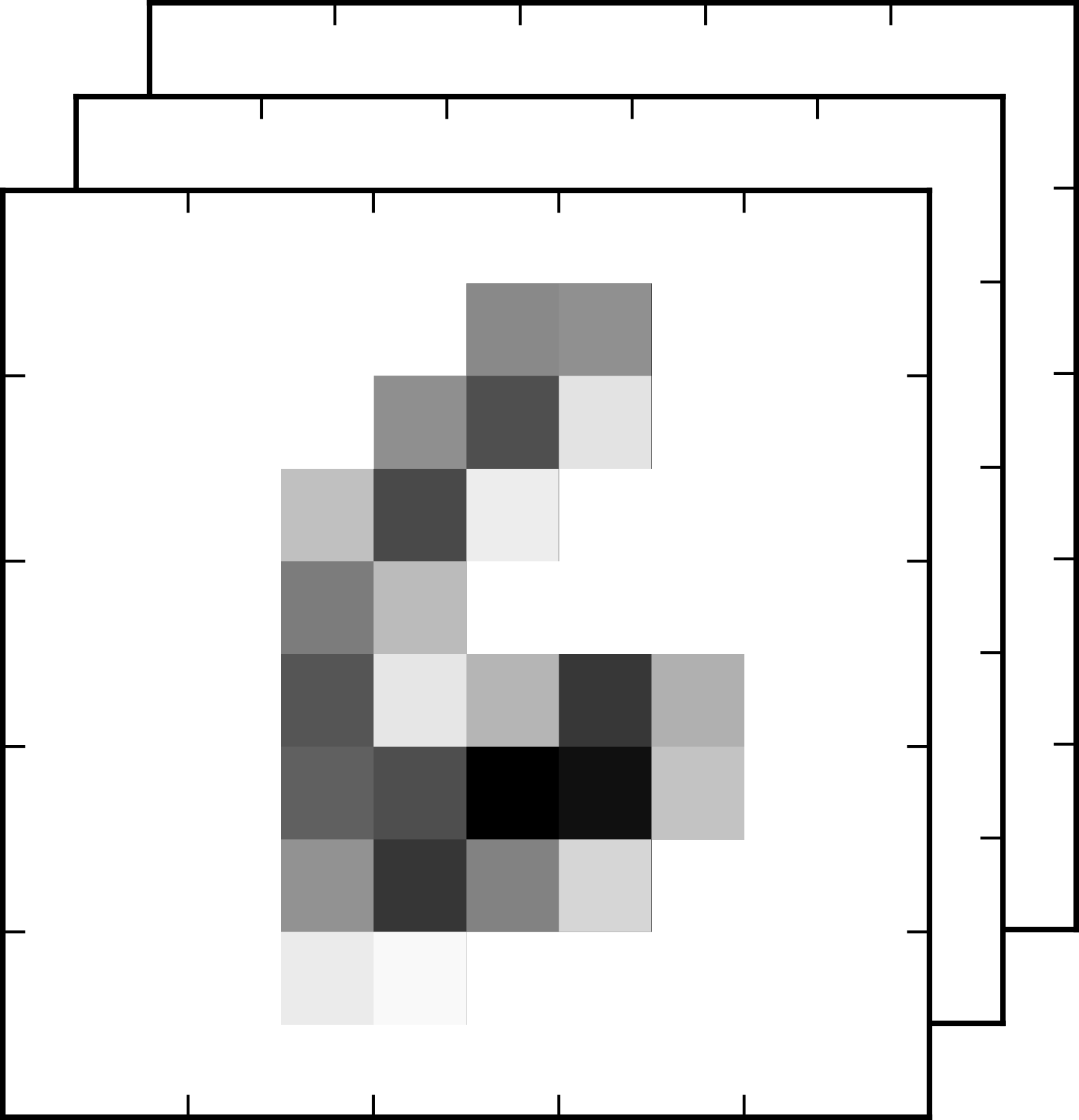}}
      \put(15,376){\includegraphics[width=0.9cm]{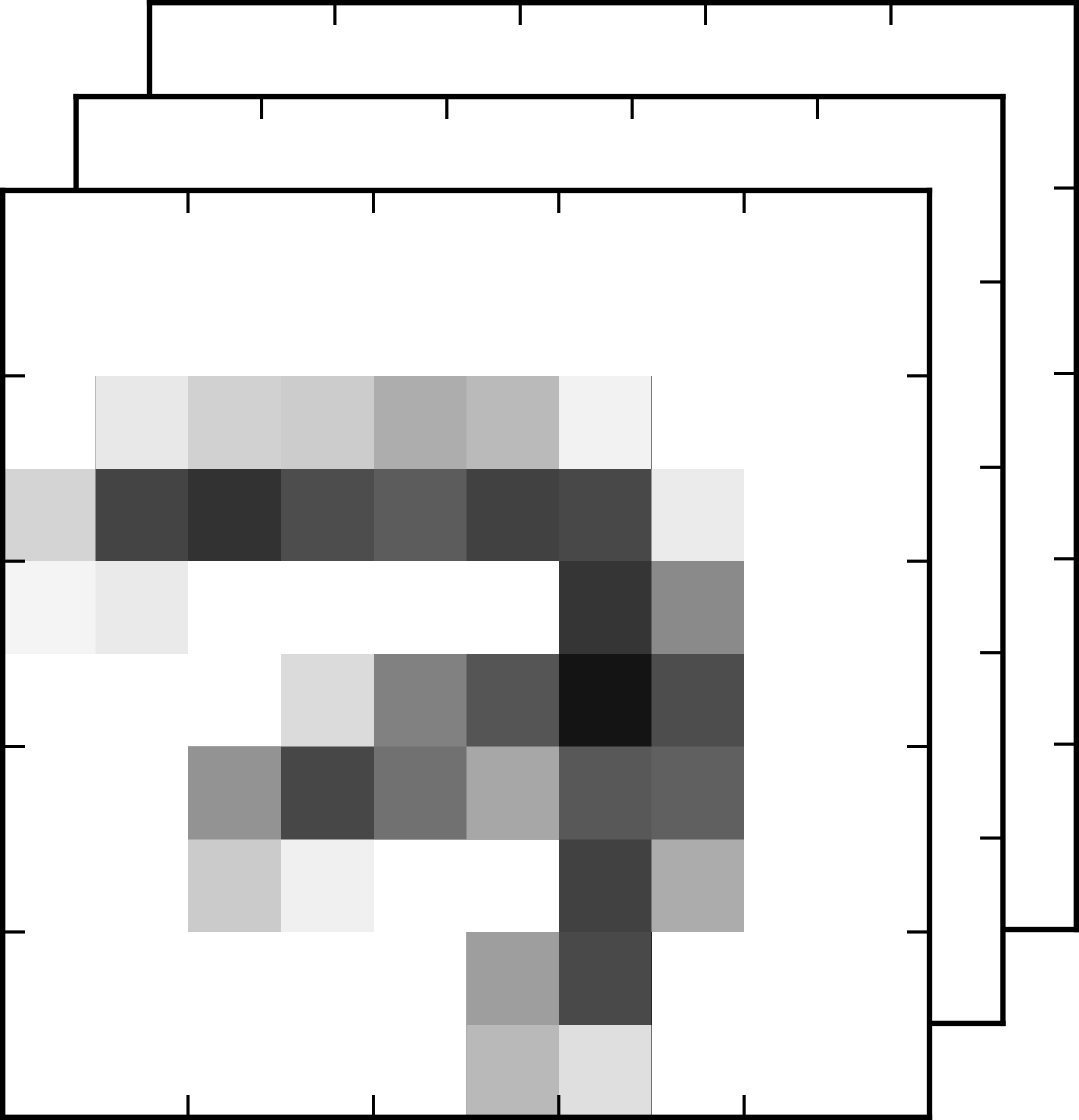}}
      \put(842,441){\includegraphics[width=0.4cm]{img/mnist_grid/mnist_grid_noopaq_0.png}}
      \put(870,441){\includegraphics[width=0.4cm]{img/mnist_grid/mnist_grid_noopaq_1.png}}
      \put(898,441){\includegraphics[width=0.4cm]{img/mnist_grid/mnist_grid_noopaq_4.png}}
      \put(926,441){\includegraphics[width=0.4cm]{img/mnist_grid/mnist_grid_noopaq_6.png}}
      \put(954,441){\includegraphics[width=0.4cm]{img/mnist_grid/mnist_grid_noopaq_7.png}}
    \end{overpic}%
  \caption{%
    Spike raster plot of the neural activity of all layers on the neuromorphic hardware after \itl/ training.
    Each horizontal dash denotes the time at which a certain neuron spiked.
    Five examples per digit are presented where in the plot same digits are denoted by the same background color.
    Correctly classified images are marked with a green circle.
  }
  \label{fig:raster_itl}
\end{figure*}

An example for the activity on the neuromorphic hardware during classification after \itl/ training for one choice of hardware neurons and initial software parameters can be found in \cref{fig:raster_itl}.
The figure shows the spike times of all neurons in the network for five presented samples of every digit.
An image is considered to have been classified correctly if the neuron associated with the input digit shows the highest activity of all label neurons.
After training, all images are correctly classified, except for the first example of digit \class{6} which was mistaken for a \class{4}.
Comparing the weights before and after \itl/ training, see \cref{fig:mnist_weight_migration}, shows that only slight adjustments are needed to compensate for hardware effects.

The evolution of the accuracy per training batch for both the software model and the \itl/ training of the hardware is shown in \cref{fig:mnist_accuracy} for \num{130} different sets of hardware neurons and initial weights of the software model.
The total classification accuracy is computed as the sum of correctly classified patterns divided by the total number of patterns in the test set.
After \num{15000} training steps, the accuracy of the software model is \SI{97}{\percent} with a negligible uncertainty arising from the choice of initial weights.
Directly after converting the \TFnet/ to the network of spiking neurons, the accuracy is reduced to \asymunc{72}{12}{10}{\si{\percent}}.
It increases to \asymunc{95}{1}{2}{\si{\percent}} at the end of the \itl/ training, being close to the performance of the software model with the uncertainty given by the interquartile range (IQR).

\section{Discussion}

For problems involving spatial pattern recognition, deep neural networks have become state of the art.
Almost by definition, they should lend themselves to implementation in neuromorphic substrates.
However, two non-trivial problems exist.
First, the input-output relationship of the abstract units used in typical deep networks needs to be mapped to spiking neuron dynamics.
Second, in case of analog hardware, distortions in these dynamics need to be take into account.
The latter is especially problematic because the performance of the network usually relies on precise parameter training.

Here, we have addressed these problems in the context of the \BSSWSS/, an accelerated analog neuromorphic platform that emulates biologically inspired neuron models.
For mapping activities from the abstract domain to spikes, we have used a rate-coding scheme.
The translation of the network topology, including connectivity structure and parameters, was described in \cref{subsec:nm_impl}.
Following this mapping of a pretrained network to the hardware substrate, the resulting distortions in dynamics and parameters have been compensated by in-the-loop training, as described in \cref{subsec:itl_training}.

This two-stage approach was evaluated for a small network trained on handwritten digits.
In this exemplary scenario, it was possible to almost completely restore the performance of the software-simulated abstract model in hardware.
An implicit, but essential component of our methodology is the fact that the backpropagation of errors needs not be precise: computing the cost function gradients using a ReLU activation function is sufficient for adapting the weights in the spiking network.
This circumvents the difficulty of otherwise having to determine an exact derivative of the cost function with respect to the LIF activation function, which would be further exacerbated by the diversity of neuronal activation functions on the analog substrate.

Here, the mapping-induced distortions in network dynamics and configuration parameters have been compensated by additional training.
A complementary approach would be to modify the network in a way that makes it more robust to hardware-induced distortions, as discussed, e.g., in~\cite{petrovici2014characterization}.
While rate-based approaches such as ours are inherently robust against jitter in the timing of spikes, robust architectures become particularly important in single-spike coding schemes, as discussed in~\cite{petrovici2016robustness}.

The proof-of-principle experiments presented here were part of the commissioning phase of the BrainScaleS system and lay the groundwork for more extensive studies.
The most interesting question to be addressed next is whether the results achieved here also hold for larger networks that can deal with more complex datasets.
Once fully functional, our system will be able to accomodate such large networks without any scaling-induced reduction in processing time due to its inherently parallel nature.

In the long run, the potentially most rewarding challenge will be to fully port the training to the hardware as well.
To this end, an integrated plasticity processor~\cite{githubnux} has been designed that will allow the emulation of different learning rules at runtime~\cite{friedmannschemmel2016}.
Learning can then also profit from the acceleration that, for now, only benefits the operation of the fully trained network.
The use of analog spiking hardware might then not only allow accelerated data processing with pre-specified networks, but also facilitate fast training of biologically inspired architectures that can, in certain contexts, even outperform classical machine learning algorithms~\cite{leng2016spiking}.

\section*{Acknowledgment}

This work has received funding from the European Union Sixth Framework Programme ([FP6/2002-2006]) under grant agreement no 15879 (FACETS), the European Union Seventh Framework Programme ([FP7/2007-2013]) under grant agreement no 604102 (HBP), 269921 (BrainScaleS) and 243914 (Brain-i-Nets) and the Horizon 2020 Framework Programme ([H2020/2014-2020]) under grant agreement no 720270 (HBP) as well as the Manfred St\"ark Foundation.

The authors wish to thank
Simon Friedmann,
Matthias Hock,
Ioannis Kokkinos,
Tobias Nonnenmacher,
Lukas Pilz,
Moritz Schilling,
Dominik Schmidt,
Sven Schrader,
Simon Ziegler,
and
Holger Zoglauer
for their contributions to the development and commissioning of the system,
Würth Elektronik GmbH \& Co.~KG in Schopfheim for the development and manufacturing of the special wafer-carrier PCB used in the \BSSWSS/,
and Fraunhofer-Institut für Zuverlässigkeit und Mikrointegration (IZM), Berlin, Germany for developing the post-processing technique which is required for wafer-wide communication and external connectivity to the wafer.

The first two authors contributed equally to this work.

\IEEEtriggeratref{21}
\bibliographystyle{IEEEtran}

\bibliography{bib/intheloop_ijcnn2017}

\begin{thebibliography}{10}
\providecommand{\url}[1]{#1}
\csname url@samestyle\endcsname
\providecommand{\newblock}{\relax}
\providecommand{\bibinfo}[2]{#2}
\providecommand{\BIBentrySTDinterwordspacing}{\spaceskip=0pt\relax}
\providecommand{\BIBentryALTinterwordstretchfactor}{4}
\providecommand{\BIBentryALTinterwordspacing}{\spaceskip=\fontdimen2\font plus
\BIBentryALTinterwordstretchfactor\fontdimen3\font minus
  \fontdimen4\font\relax}
\providecommand{\BIBforeignlanguage}[2]{{%
\expandafter\ifx\csname l@#1\endcsname\relax
\typeout{** WARNING: IEEEtran.bst: No hyphenation pattern has been}%
\typeout{** loaded for the language `#1'. Using the pattern for}%
\typeout{** the default language instead.}%
\else
\language=\csname l@#1\endcsname
\fi
#2}}
\providecommand{\BIBdecl}{\relax}
\BIBdecl

\bibitem{LeCun2015}
Y.~LeCun, Y.~Bengio, and G.~Hinton, ``Deep learning,'' \emph{Nature}, vol. 521,
  no. 7553, pp. 436--444, May 2015.

\bibitem{furber2016-large-scale-nm-system}
S.~Furber, ``Large-scale neuromorphic computing systems,'' \emph{J Neural Eng},
  vol.~13, no.~5, p. 051001, 2016.

\bibitem{maass1997fast}
W.~Maass, ``Fast sigmoidal networks via spiking neurons,'' \emph{Neural
  Comput}, vol.~9, no.~2, pp. 279--304, 1997.

\bibitem{Esser20092016}
S.~K. Esser, P.~A. Merolla, J.~V. Arthur \emph{et~al.}, ``Convolutional
  networks for fast, energy-efficient neuromorphic computing,'' \emph{Proc.
  Natl. Acad. Sci. U.S.A.}, 2016.

\bibitem{hohmann_ijcnn04}
S.~G. Hohmann, J.~Fieres, K.~Meier \emph{et~al.}, ``Training fast mixed-signal
  neural networks for data classification,'' in \emph{Proc Int Jt Conf Neural
  Netw}, vol.~4.\hskip 1em plus 0.5em minus 0.4em\relax IEEE Press, Jul. 2004,
  pp. 2647--2652.

\bibitem{fieres06convolutional}
J.~Fieres, J.~Schemmel, and K.~Meier, ``A convolutional neural network tolerant
  of synaptic faults for low-power analog hardware,'' in \emph{Proceedings of
  2nd IAPR International Workshop on Artificial Neural Networks in Pattern
  Recognition}, ser. Springer Lecture Notes in Artificial Intelligence, vol.
  4087.\hskip 1em plus 0.5em minus 0.4em\relax Ulm, Germany: Springer
  International Publishing, Aug. 2006, pp. 122--132.

\bibitem{pfeil2013six}
T.~Pfeil, A.~Gr\"ubl, S.~Jeltsch \emph{et~al.}, ``Six networks on a universal
  neuromorphic computing substrate,'' \emph{Frontiers in Neuroscience}, vol.~7,
  p.~11, 2013.

\bibitem{liu2015event}
S.-C. Liu, \emph{Event-based neuromorphic systems}.\hskip 1em plus 0.5em minus
  0.4em\relax John Wiley \& Sons, 2015.

\bibitem{schemmeliscas2010}
J.~Schemmel, D.~Br\"uderle, A.~Gr\"ubl \emph{et~al.}, ``A wafer-scale
  neuromorphic hardware system for large-scale neural modeling,'' in \emph{IEEE
  Int Symp Circuits Syst Proc}, May 2010, pp. 1947--1950.

\bibitem{brette2005adex}
R.~Brette and W.~Gerstner, ``Adaptive exponential integrate-and-fire model as
  an effective description of neuronal activity,'' \emph{J. Neurophysiol.},
  vol.~94, no.~5, pp. 3637--3642, 2005.

\bibitem{millner2010AdEx}
S.~Millner, A.~Gr\"{u}bl, K.~Meier \emph{et~al.}, ``A {VLSI} implementation of
  the adaptive exponential integrate-and-fire neuron model,'' in \emph{Adv Neur
  In}, J.~Lafferty, C.~K.~I. Williams, J.~Shawe-Taylor \emph{et~al.}, Eds.,
  vol.~23, 2010, pp. 1642--1650.

\bibitem{naud2008firing}
\BIBentryALTinterwordspacing
R.~Naud, N.~Marcille, C.~Clopath \emph{et~al.}, ``Firing patterns in the
  adaptive exponential integrate-and-fire model,'' \emph{Biological
  Cybernetics}, vol.~99, no.~4, pp. 335--347, Nov 2008. [Online]. Available:
  \url{http://dx.doi.org/10.1007/s00422-008-0264-7}
\BIBentrySTDinterwordspacing

\bibitem{schemmel_ijcnn2008}
J.~Schemmel, J.~Fieres, and K.~Meier, ``Wafer-scale integration of analog
  neural networks,'' in \emph{Proc Int Jt Conf Neural Netw}, Hong Kong, Jul.
  2008.

\bibitem{thanasoulis2014pulse}
V.~Thanasoulis, B.~Vogginger, J.~Partzsch \emph{et~al.}, ``A pulse
  communication flow ready for accelerated neuromorphic experiments,'' in
  \emph{IEEE Int Symp Circuits Syst Proc}, Jun. 2014, pp. 265--268.

\bibitem{scholze2011vlsi}
S.~Scholze, S.~Schiefer, J.~Partzsch \emph{et~al.}, ``{VLSI} implementation of
  a 2.8 {GEvent/s} packet-based {AER} interface with routing and event sorting
  functionality,'' \emph{Front Neurosci}, vol.~5, p. 117, 2011.

\bibitem{petrovici2014characterization}
M.~A. Petrovici, B.~Vogginger, P.~M{\"u}ller \emph{et~al.}, ``Characterization
  and compensation of network-level anomalies in mixed-signal neuromorphic
  modeling platforms,'' \emph{PLoS ONE}, vol.~9, no.~10, p. e108590, 2014.

\bibitem{raikov2011nineml}
I.~Raikov, R.~Cannon, R.~Clewley \emph{et~al.}, ``{NineML}: the network
  interchange for neuroscience modeling language,'' \emph{BMC Neuroscience},
  vol.~12, no.~1, p. P330, 2011.

\bibitem{gleeson2010neuroml}
P.~Gleeson, S.~Crook, R.~C. Cannon \emph{et~al.}, ``{NeuroML}: A language for
  describing data driven models of neurons and networks with a high degree of
  biological detail,'' \emph{PLoS Comput Biol}, vol.~6, no.~6, p. e1000815,
  Jun. 2010.

\bibitem{davison08pynn}
A.~P. Davison, D.~Br\"{u}derle, J.~Eppler \emph{et~al.}, ``{PyNN}: a common
  interface for neuronal network simulators,'' \emph{Front Neuroinform},
  vol.~2, no.~11, 2008.

\bibitem{bruederle09establishing}
D.~Br{\"u}derle, E.~M{\"u}ller, A.~Davison \emph{et~al.}, ``Establishing a
  novel modeling tool: A python-based interface for a neuromorphic hardware
  system,'' \emph{Front Neuroinform}, vol.~3, no.~17, 2009.

\bibitem{djurfeldt2012csa}
M.~Djurfeldt, ``The connection-set algebra---a novel formalism for the
  representation of connectivity structure in neuronal network models,''
  \emph{Neuroinformatics}, vol.~10, no.~3, pp. 287--304, 2012.

\bibitem{denker2015elephant}
M.~Denker, A.~Yegenoglu, D.~Holstein \emph{et~al.}, ``elephant: {A}n
  open-source tool for the analysis of electrophysiological data.'' in
  \emph{Proceedings of the 11th Meeting of the German Neuroscience Society,
  Neuroforum 2015}.\hskip 1em plus 0.5em minus 0.4em\relax German Neuroscience
  Society, Mar 2015, pp. T27--2B.

\bibitem{garcia2014neo}
S.~Garcia, D.~Guarino, F.~Jaillet \emph{et~al.}, ``Neo: an object model for
  handling electrophysiology data in multiple formats,'' \emph{Front
  Neuroinform}, vol. 8:10, February 2014.

\bibitem{jette2003slurm}
M.~Jette and M.~Grondona, ``{SLURM}: Simple linux utility for resource
  management,'' in \emph{Proceedings of ClusterWorld Conference and Expo}, San
  Jose, California, 2003.

\bibitem{Gewaltig:NEST}
M.-O. Gewaltig and M.~Diesmann, ``{NEST} ({NE}ural {S}imulation {T}ool),''
  \emph{Scholarpedia}, vol.~2, no.~4, p. 1430, 2007.

\bibitem{lecunmnist}
\BIBentryALTinterwordspacing
Y.~LeCun and C.~Cortes, ``The {MNIST} database of handwritten digits,'' 1998.
  [Online]. Available: \url{http://yann.lecun.com/exdb/mnist}
\BIBentrySTDinterwordspacing

\bibitem{Cao2015}
Y.~Cao, Y.~Chen, and D.~Khosla, ``Spiking deep convolutional neural networks
  for energy-efficient object recognition,'' \emph{Int J Comput Vis}, vol. 113,
  no.~1, pp. 54--66, 2015.

\bibitem{tensorflow2015-whitepaper}
\BIBentryALTinterwordspacing
M.~Abadi, A.~Agarwal, P.~Barham \emph{et~al.}, ``{TensorFlow}: Large-scale
  machine learning on heterogeneous systems,'' Google Research, Whitepaper,
  2015, software available from tensorflow.org. [Online]. Available:
  \url{http://tensorflow.org/}
\BIBentrySTDinterwordspacing

\bibitem{Qian1999145}
N.~Qian, ``On the momentum term in gradient descent learning algorithms,''
  \emph{Neural Netw.}, vol.~12, no.~1, pp. 145--151, 1999.

\bibitem{10.3389/fnins.2013.00178}
P.~O'Connor, D.~Neil, S.-C. Liu \emph{et~al.}, ``Real-time classification and
  sensor fusion with a spiking deep belief network,'' \emph{Front Neurosci},
  vol.~7, p. 178, 2013.

\bibitem{petrovici2016robustness}
M.~A. Petrovici, A.~Schroeder, O.~Breitwieser \emph{et~al.}, ``Robustness from
  structure: fast inference on a neuromorphic device with hierarchical {LIF}
  networks,'' submitted to IJCNN 2017, 2016.

\bibitem{githubnux}
\BIBentryALTinterwordspacing
S.~Friedmann, ``The {Nux} processor v3.0,'' Electronic Vision(s) Group,
  Kirchhoff-Institute for Physics, Heidelberg University, User Guide, 2015.
  [Online]. Available: \url{https://github.com/electronicvisions/nux}
\BIBentrySTDinterwordspacing

\bibitem{friedmannschemmel2016}
S.~Friedmann, J.~Schemmel, A.~Gr\"ubl \emph{et~al.}, ``Demonstrating hybrid
  learning in a flexible neuromorphic hardware system,'' \emph{IEEE Trans.
  Biomed. Circuits Syst.}, vol.~PP, no.~99, pp. 1--15, 2016.

\bibitem{leng2016spiking}
L.~Leng, M.~A. Petrovici, R.~Martel \emph{et~al.}, ``Spiking neural networks as
  superior generative and discriminative models,'' in \emph{Cosyne Abstracts,
  Salt Lake City USA}, February 2016.

\end{thebibliography}

\end{document}